\documentclass[journal]{IEEEtran}
\ifCLASSINFOpdf
\else
\fi
\usepackage{amsmath}
\usepackage{makecell}
\usepackage{booktabs}
\usepackage{multirow}
\usepackage{algorithm}
\usepackage{algpseudocode}
\usepackage{graphicx}
\usepackage[caption=false,font=footnotesize]{subfig}
\usepackage{xcolor}
\usepackage[colorlinks=true,
            linkcolor=blue,
            citecolor=blue,
            filecolor=blue,
            urlcolor=blue]{hyperref}

\begin{document}
\title{MG-HGNN: A Heterogeneous GNN Framework\\ for Indoor Wi-Fi Fingerprint-Based Localization}
\author{Yibu Wang,
        Zhaoxin Zhang,
        Ning Li,~\IEEEmembership{Member,~IEEE,}
        Xinlong Zhao,
        Dong Zhao,
        Tianzi Zhao
\thanks{\textbf{This work has been submitted to the IEEE for possible publication. Copyright may be transferred without notice, after which this version may no longer be accessible.} This work is supported by the National Key Research and Development Program of China [Grant No. 2024QY1103], the Shandong Provincial Natural Science Foundation, China [Grant No. ZR2024QF138].\textit{(Corresponding author: Zhaoxin Zhang.)}}
\thanks{Yibu Wang, Zhaoxin Zhang, Ning Li, and Tianzi Zhao are with the School of Computer Science and Technology, Harbin Institute of Technology, China (e-mail: 24b903081@stu.hit.edu.cn; heart@hit.edu.cn; li.ning@hit.edu.cn; 23b903088@stu.hit.edu.cn).}
\thanks{Xinlong Zhao is with the China Mineral Resources Group Big Data Co., Ltd, China (e-mail: xinlong.zhao@qq.com).}
\thanks{Dong Zhao is with the Tianhe Cyberspace Security Technology Research Institute Co., China (e-mail: zhaod@skyvault.cn).}}

\maketitle

\begin{abstract}
Received signal strength indicator (RSSI) is the primary representation of Wi-Fi fingerprints and serves as a crucial tool for indoor localization. However, existing RSSI-based positioning methods often suffer from reduced accuracy due to environmental complexity and challenges in processing multi-source information. To address these issues, we propose a novel multi-graph heterogeneous GNN framework (MG-HGNN) to enhance spatial awareness and improve positioning performance. In this framework, two graph construction branches perform node and edge embedding, respectively, to generate informative graphs. Subsequently, a heterogeneous graph neural network is employed for graph representation learning, enabling accurate positioning. The MG-HGNN framework introduces the following key innovations: 1) multi-type task-directed graph construction that combines label estimation and feature encoding for richer graph information; 2) a heterogeneous GNN structure that enhances the performance of conventional GNN models. Evaluations on the UJIIndoorLoc and UTSIndoorLoc public datasets demonstrate that MG-HGNN not only achieves superior performance compared to several state-of-the-art methods, but also provides a novel perspective for enhancing GNN-based localization methods. Ablation studies further confirm the rationality and effectiveness of the proposed framework.
\end{abstract}

\begin{IEEEkeywords}
Fingerprint-based localization, graph neural network, heterogeneous network, received signal strength indicator (RSSI).
\end{IEEEkeywords}

% \fi
\IEEEpeerreviewmaketitle

\section{Introduction}
\label{sec:introduction}
\IEEEPARstart{I}{ndoor} localization technologies aim to estimate the position of mobile users or devices in indoor environments where satellite-based systems such as GPS are ineffective~\cite{zafari2019survey}. Over the past decade, a variety of wireless indoor localization techniques have been developed based on different sensing modalities, including Bluetooth Low Energy (BLE)~\cite{9517109}, Ultra Wideband (UWB)~\cite{10660592}, Radio Frequency Identification (RFID)~\cite{6404550}, magnetic field sensing~\cite{shu2015magicol}, and Wi-Fi~\cite{abbas2019wideep},~\cite{chen2019wifi}. Among them, Wi-Fi based localization has attracted a lot of attention due to the ubiquity of Wi-Fi infrastructure, low deployment cost, and compatibility with existing mobile devices without requiring additional hardware~\cite{zafari2019survey}.

Wi-Fi based localization techniques can generally be divided into two categories: geometric localization and fingerprint-based localization~\cite{zafari2019survey}. Geometric approaches \cite{neri2010doa},~\cite{li2018tdoa},~\cite{zheng2018exploiting},~\cite{ilci2015trilateration} rely on physical signal propagation models and geometric relationships between access points (AP) and target devices. Although these methods achieve high accuracy under ideal conditions, their performance could be affected by multipath propagation and environmental dynamics, which are ubiquitous in complex indoor spaces. In contrast, fingerprint-based localization does not depend on strict geometric models, it utilizes radio frequency (RF) signal fingerprints measured at different locations to train data-driven mappings between signal features and positions~\cite{he2015wi}. This paradigm has shown stronger adaptability and robustness to environmental variations, leading to its widespread adoption in practical indoor localization systems.

A Wi-Fi fingerprint localization system typically consists of two phases: offline training and online positioning. During the offline phase, the fingerprints of received signal strength indicator (RSSI) or channel state information (CSI) are collected at reference points to build a fingerprint database, and training process is conducted using it. During the online phase, the system uses trained mappings and the real-time fingerprint to estimate the user's position. Although CSI can provide fine-grained multipath information and higher theoretical accuracy, it requires specialized hardware and introduces significant complexity in data acquisition. Therefore, real-world systems and public datasets~\cite{lohan2017wi},~\cite{torres2014ujiindoorloc},~\cite{song2019novel} primarily use RSSI-based fingerprints due to their accessibility, ease of collection, and hardware compatibility.

Existing RSSI-based localization methods can be broadly grouped into two families: traditional machine learning (ML) approaches and deep learning (DL) based methods. Traditional machine learning approaches have achieved satisfactory performance in small-scale or static indoor environments~\cite{ge2016optimization},~\cite{8485369},~\cite{9509849}; however, their dependence on feature engineering and the difficulty in adapting to complex environments may limit their scalability and robustness. These limitations motivate the adoption of deep learning based methods, including various neural network architectures such as convolutional networks, autoencoders, and generative models~\cite{qin2021ccpos},~\cite{njima2021indoor}. While deep learning models enable non-linear mapping and high-level feature abstraction, they often process each sample independently and thus may overlook the spatial or signal correlations among samples. To address this challenge, graph neural networks (GNN)~\cite{lezama2021indoor} have been introduced. GNNs are capable of using graph structures to capture complex internal dependencies in signal and location spaces, thus achieving improved performance. 

Currently, GNN-based localization methods present several challenges. Existing graph construction strategies primarily rely on signal feature similarity, which may not always reflect the true spatial proximity among fingerprint samples, potentially leading to suboptimal graph structures. Such discrepancies between the graph topology and the physical layout can affect the effectiveness of feature aggregation and the positioning accuracy. Furthermore, most approaches adopt homogeneous graph representations, which may not fully capture the heterogeneous and multi-relational dependencies inherently present in fingerprint data. These limitations highlight the need for more flexible and expressive graph representations to enhance model performance in complex indoor environments.

In this article, to address the challenges above, we propose a novel multi-graph heterogeneous GNN framework (MG-HGNN) for Wi-Fi fingerprint-based localization, including task-directed multi-graph construction, heterogeneous graph neural network based localization and other modules. This framework can make full use of RSSI fingerprint information, and utilize multiple graph structures to capture a variety of task-related information. At the same time, inductive learning is implemented to improve the generalization capability. This framework can also serve as a design paradigm for enhancing existing GNN models. Our main contributions are as follows.

\hangindent=2.2em
\hangafter=1
1) We proposed a conceptual framework of heterogeneous GNN-based fingerprint localization, and further developed a practical Wi-Fi RSSI fingerprint localization framework built upon this concept.

\hangindent=2.2em
\hangafter=1
2) We design a multi-type task-directed graph construction structure consisting of a GNN based localization model and a stacked feature encoder (SFE). This structure can produce informative position-based and feature-based graphs for the heterogeneous model.

\hangindent=2.2em
\hangafter=1
3) We present a heterogeneous GNN model as the key part of the proposed localization framework, this model utilizes the multi-type graphs to perform positioning. The overall framework is validated and tested on two public datasets, and the experimental results show that our framework outperforms state-of-the-art techniques.

The remainder of this article is organized as follows. Section~\ref{sec:relatedwork} is related work on Wi-Fi fingerprint-based localization, Section~\ref{sec:overview} is an overview of the localization framework we proposed, Section~\ref{sec:MG-HGNN} is the detailed introduction to each module of the framework, Section~\ref{sec:experiment} provides the experimental results and analysis, and Section~\ref{sec:conclusion} is our final conclusion and future work.

\section{Related Work}
\label{sec:relatedwork}

\subsection{Traditional ML Based Approaches}
Traditional Wi-Fi fingerprint localization methods primarily rely on classical machine learning algorithms to establish a mapping between received signal strength (RSS) values and spatial coordinates. These approaches typically assume that fingerprints collected at different locations can be modeled as feature–label pairs and learned through supervised regression or classification frameworks.

Early studies introduced algorithms such as KNN and SVM to predict the user’s position. For instance, Bi et al. presented an adaptive weighted KNN positioning method based on an omnidirectional fingerprint database and twice affinity propagation clustering~\cite{bi2018adaptive}, and Hu et al. proposed a self-adaptive WKNN (SAWKNN) algorithm with a dynamic K~\cite{hu2018experimental}. These two KNN methods focus on improving adaptive neighbor selection strategies to enhance accuracy and robustness. In another study, Zhang et al. introduced user posture incorporating to SVM to achieve lower localization errors~\cite{zhang2019improving}. More recently, Umair et al. formulated a robust Wi-Fi fingerprinting approach based on geospatial cells to realize generic and robust positioning~\cite{umair2025robust}, providing a new way to enhance KNN-based localization methods.

\subsection{Deep Learning Based Localization}
With the advancement of neural network technologies, deep learning techniques have been widely adopted to enhance the representation and generalization capability of Wi-Fi fingerprint localization models. Deep learning based methods can automatically learn hierarchical feature representations from large-scale RSSI data. Recently, a variety of neural network architectures have been explored, including multilayer perceptrons (MLP), convolutional neural networks (CNN), and auto-encoders (AE). For example, Song et al. proposed CNNLoc, a localization system combining a stacked auto-encoder (SAE) with a one-dimensional CNN~\cite{song2019novel}, Cha et al. proposed a hierarchical auxiliary deep neural network (HADNN) to reduce hierarchical information error~\cite{cha2022hierarchical}. Alitaleshi et al. introduced an indoor positioning method based on extreme learning machine autoencoder (ELM-AE) and two-dimensional CNN~\cite{alitaleshi2023ea}. More recently, Ayinla et al. proposed a Wi-Fi indoor localization framework using recursive feature elimination with cross-validation (RFECV) and deep neural networks with batch normalization (DNNBN)~\cite{ayinla2024enhanced}, and Wu et al. proposed LCVAE-CNN, which integrates a location-conditioned variational autoencoder (LCVAE) and a multitask CNN to perform data argumentation and positioning~\cite{wu2025lcvae}. Although deep learning approaches have achieved remarkable progress, they often process fingerprints as independent observations, without explicitly modeling the intrinsic relationships among them. This may limit the model’s ability to capture spatial continuity and contextual correlations essential for accurate localization.

\subsection{Graph Neural Network Based Fingerprinting Methods}
In recent years, graph neural networks have emerged as a powerful tool for modeling relational data and have been increasingly applied to Wi-Fi fingerprint localization. By representing each fingerprint as a node and defining edges based on spatial, semantic or signal relationships, GNN can explicitly capture the topological structure underlying indoor environments. Researchers have employed GNNs, such as graph convolutional networks (GCN)~\cite{kipf2016semi}, graph sample and aggregate (GraphSAGE)~\cite{hamilton2017inductive} and graph attention network (GAT)~\cite{velickovic2017graph} to aggregate information from neighboring nodes and perform graph representation learning, thus obtaining improved localization performance. Zhang et al. presented a Wi-Fi domain adversarial graph convolutional network model to enhance positioning accuracy using crowdsensed data~\cite{zhang2023domain}. Zheng et al. proposed GraFin, a applicable graph-based fingerprinting approach for accurate and robust indoor positioning~\cite{zheng2021grafin}. Moreover, Luo and Meratnia introduced a geometric deep learning framework for indoor localization~\cite{luo2022geometric}. This method utilizes homogeneous and heterogeneous GraphSAGE architectures to learn from AP based subgraphs. However, its dependency on AP location information may constrain its use in practical Wi-Fi fingerprinting scenarios, where such information is often unknown. In other recent studies, Wang et al. proposed a localization system with GNN and access point selection~\cite{wang2024graph}, and Zhang et al. proposed a fingerprint-based localization methods using residual graph attention network (Res-GAT)~\cite{zhang2025graphloc}. These two studies introduced graph attention networks to obtain better aggregation and node representation.

GNN-based methods have shown remarkable improvements in positioning accuracy and robustness; however, several important challenges remain. Many of these existing GNN-based approaches depend on known or pre-estimated AP locations, which are often unavailable in practical scenarios. As a result, centroid-based approximations are commonly adopted, but may introduce structural bias that can adversely affect positioning accuracy. Moreover, most existing methods construct graphs using a single similarity metric, which limits their capacity to represent complex spatial relationships and may lead to distorted graph structures that increase localization errors. In addition, the naturally existing multi-type relationship information in indoor environments is often overlooked, reducing the models’ ability to capture heterogeneous spatial dependencies and thereby constraining overall robustness. To overcome these challenges, this paper proposes a heterogeneous GNN framework that utilizes multi-type graph structures and inductive learning to producing improved localization results with better generalization and environmental adaptive capability.

\begin{figure*}
    \centering
    \includegraphics[width=\linewidth]{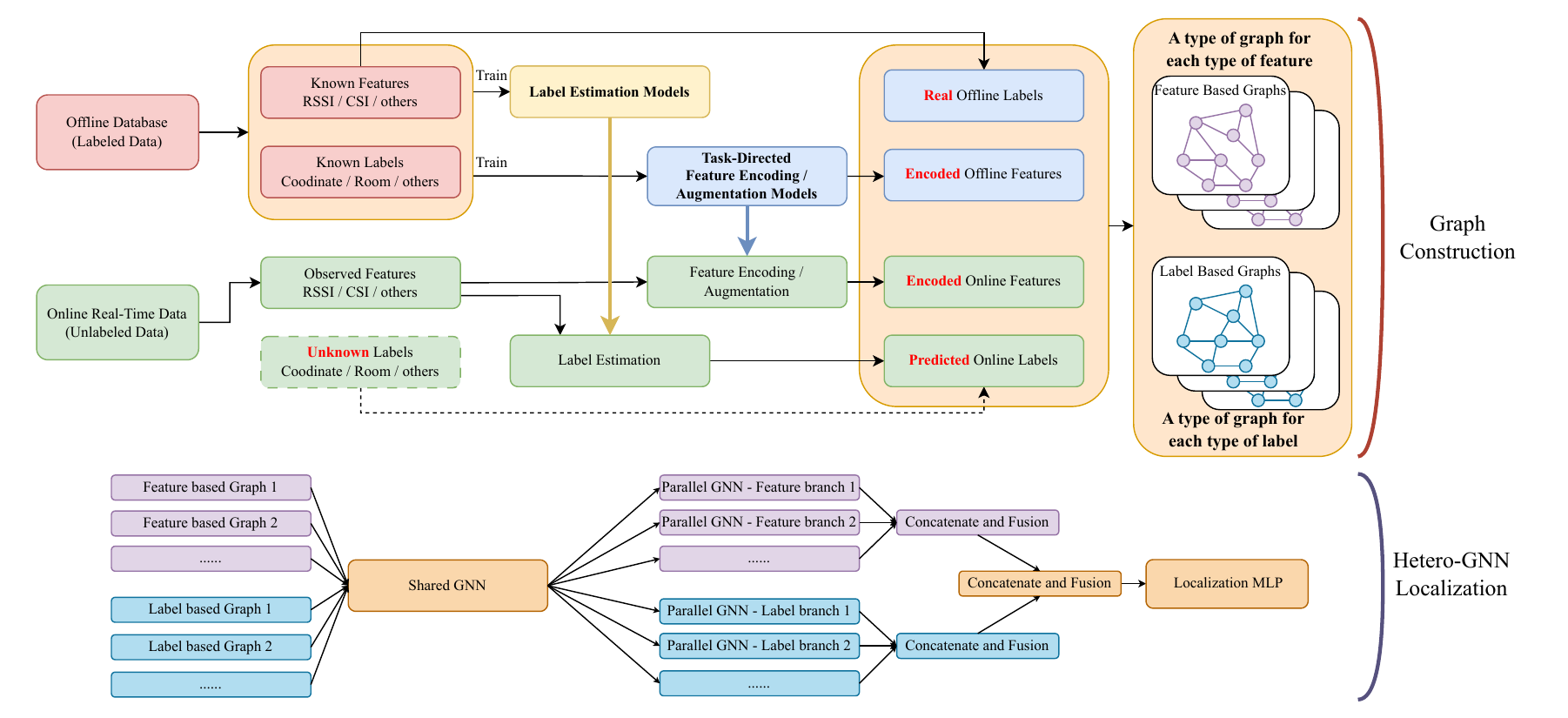}
    \caption{Conceptual framework of heterogeneous GNN based fingerprint localization.}
    \label{fig:3.1}
\end{figure*}

\begin{figure*}
    \centering
    \includegraphics[width=\linewidth]{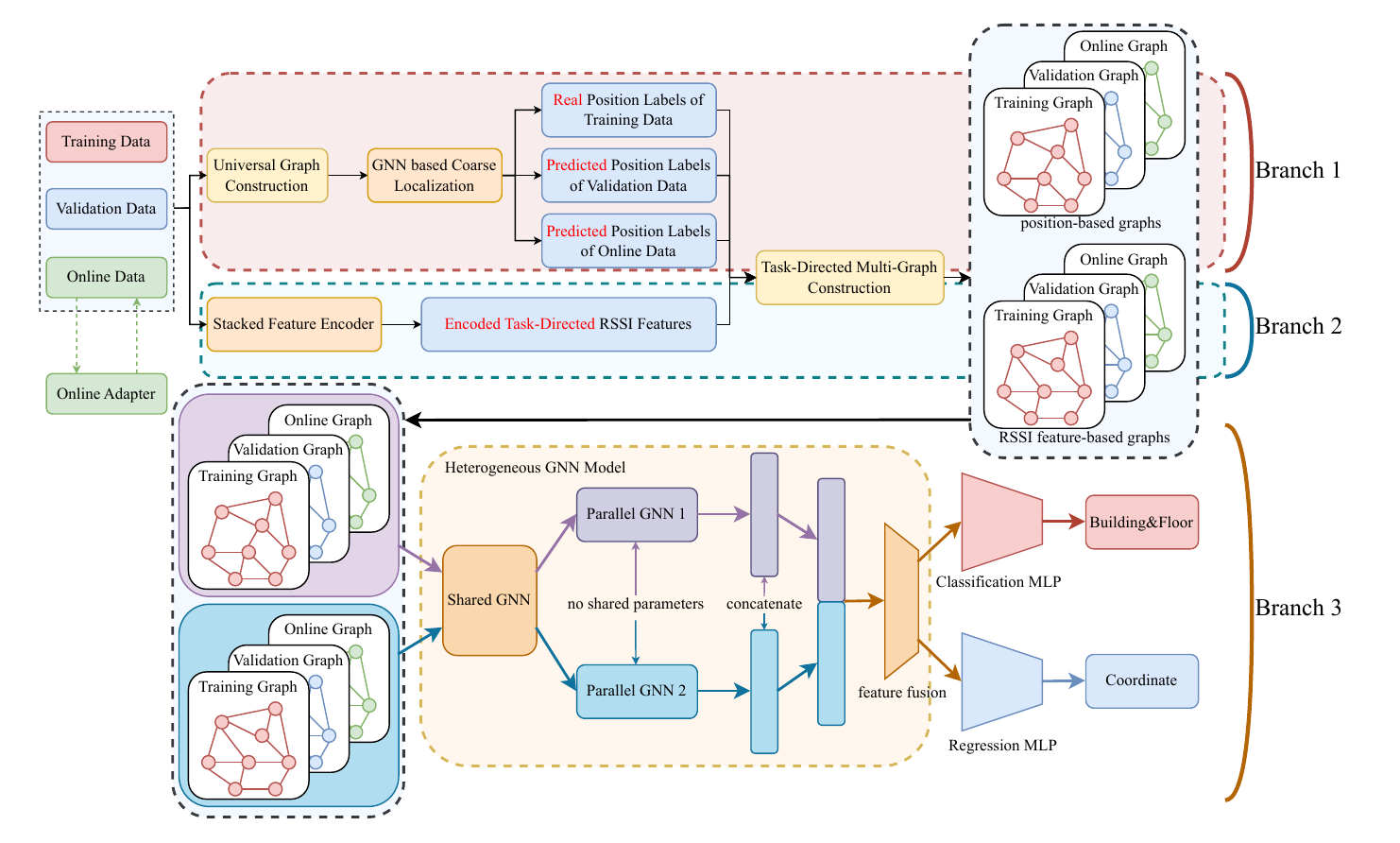}
    \caption{MG-HGNN Framework overview. Branches 1 and 2 are the graph construction branches, and branch 3 is the localization branch.}
    \label{fig:3.2}
\end{figure*}

\section{Framework Overview}
\label{sec:overview}

Fig.~\ref{fig:3.1} illustrates the proposed conceptual framework of heterogeneous GNN based fingerprint localization, which consists of two main stages: graph construction and localization. In the graph construction stage, feature-based graphs and label-based graphs are generated through task-directed feature encoding and label estimation. The localization stage employs a shared GNN layer to preliminarily process all graphs, ensuring parameter sharing and mitigating excessive feature discrepancies among different branches. Subsequently, parallel GNN layers are utilized to process each graph independently, after which their outputs are concatenated and fused to form the final graph representation. This learned representation is then fed into the downstream localization module. The MG-HGNN framework proposed in this paper is designed following this conceptual framework and is dedicated to Wi-Fi RSSI fingerprint positioning.

The overall structure of proposed localization framework is shown in Fig.~\ref{fig:3.2}. In the offline phase, the inputs are the training data and verification data, while in the online phase, the inputs are the real-time RSSI fingerprint data. The whole framework will output localization results of different granularity: building/floor/coordinate.

\subsection{Branch 1: Position-based Graph Construction}
This branch aims to construct a position-based graph structure, i.e. edge connection is based on spatial distance. We design a light-weight GNN-based coarse localization method to produce the approximate position of validation or online nodes; then we combine the predicted position labels with real position labels from the training nodes to form a position-based graph structure. This branch can be considered as the coarse part of a coarse-to-fine localization strategy; its role is to identify the approximate target region and transfer this information in the form of a graph structure. The building classification task is relatively simple; therefore, its final results will be from here and no subsequent work is required.

\subsection{Branch 2: Feature-based Graph Construction}
The purpose of this branch is to construct a task-directed RSSI feature-based graph structure, i.e. edge connection is based on feature distance using RSSI features after task-directed transformation. We design a rapid stacked feature encoder (SFE) model inspired by stacked auto-encoder, it contains encoder part for feature transformation and multilayer preceptron part for end-to-end training. This SFE model will transform origin features into more discriminative features for specific tasks, e.g. coordinates regression tasks, and finally this branch will produce a RSSI feature-based graph structure.

\subsection{Branch 3: Heterogeneous GNN based Localization}
The final localization results are produced by branch 3. We design a multi-graph heterogeneous GNN model to perform comprehensive inductive graph representation learning, and produce results using feature fusion and MLP. Two separate graph structures from branches 1 and 2 will be fed into the model, and then the localization results will be produced as final results (combining the building results from branch 1). This branch is the core part of the proposed framework, and its structure supports the use of different types of graph neural network.

\subsection{Offline and Online Work Flow}
In the offline phase, the inputs contain training and validation data. Firstly, train models in branches 1 and 2. Both branches support independent training, and we conduct separate training here. Secondly, input training and validation data into branches 1 and 2 separately to obtain two types of graph structure, and each type contains two graphs, namely training graph (only training nodes included) and validation graph(training nodes and validation nodes included) for inductive learning. Thirdly, train the model of branch 3, and save all branch parameters as localization system parameters after training and validating.

In the online phase, input the online data into branches 1 and 2 to form online graphs, and feed them into branch 3 to obtain final results. Here an online adapter (OA) can be optionally adopted: A small number of online data is input into the adaptive prepositional model for rapid training, and the adapter model is used to process input RSSI features after training. This OA is proposed to relieve the problem of data irrelevance between training and online data, it needs a part of online data for training. Therefore, we will not use it for comparison.

\section{Multi-Graph Heterogeneous GNN Framework for Fingerprint-Based Localization}
\label{sec:MG-HGNN}
In this section, we will introduce the detailed implementation of the proposed MG-HGNN framework, which contains the following parts: data preprocessing, universal graph construction, GNN-based coarse localization, stacked feature encoder, task-directed multi-graph construction, heterogeneous GNN based localization.

\subsection{Data Preprocessing}
\label{subsec:1}
Data preprocessing is an important step for data analysis and machine learning initialization. It includes data cleaning, data scaling and data transformation. For original Wi-Fi fingerprint data, the RSSI features in the data are processed first. Most Wi-Fi fingerprinting scenarios, such as UJIIndoorLoc and UTSIndoorLoc, use \textit{+100} to represent missing WAP values~\cite{torres2014ujiindoorloc},~\cite{song2019novel}, and we will start with handling missing values. Here we take the UJIIndoorLoc data set as an example: The valid RSSI feature range of UJIIndoorLoc data set is \textit{[-104, 0]}, so we replace the missing value \textit{+100} with \textit{-105}, then add \textit{105} to all features and divide the replaced RSSI features by \textit{105} after, mapping them to range \textit{[0,1]}, as shown in the following normalization formula:

\begin{equation}
    X_{\text{RSSI}} = 
    \begin{cases}
        X_{\min} - 1 & \text{if } X_{\text{RSSI}} \text{ undetected} \\
        X_{\text{RSSI}} & \text{otherwise}
    \end{cases}
\end{equation}

\begin{equation}
    X'_{\text{RSSI}} = \frac{
        X_{\text{RSSI}} - (X_{\min} - 1)
    }{
        X_{\max}-(X_{\min} - 1)
    }
\end{equation}

After processing RSSI features, we continue to process position coordinates. As an example, the position coordinates of UJIIndoorLoc data set are longitude and latitude in meters with UTM from WGS84~\cite{torres2014ujiindoorloc}. Here we use the standardization formula to process them, $\mu$ and $\sigma$ in the formula represent the mean and standard deviation of longitude and latitude, respectively:

\begin{equation}
    Y_{\text{coord\_normalized}} = \frac{
    Y_{\text{coord}} - \mu(Y_{\text{coord}})
    }{
        \sigma(Y_{\text{coord}})
    }
\end{equation}

In the implementation process, we use \textit{Standardscaler} to realize this coordinate transformation, and save its parameters after the transformation, which allows us to restore the coordinate regression output of the localization model to the actual longitude and latitude coordinate space.

For building and floor labels, we introduce multi-dimension output, \textit{softmax} (\ref{formula:4}) and \textit{argmax} in the subsequent models. Therefore, there is no need for normalization.

\begin{equation}
    \text{Softmax}(\mathbf{Pred})_j = 
    \frac{
        e^{Pred_j}
    }{
        \sum_{k=1}^{K} e^{Pred_k}
    } \quad \text{for } j = 1, 2, \dots, K
    \label{formula:4}
\end{equation}

In addition, for cases where the floor labels include underground floors, we will add an offset to make the range start from 0. For instance, in the UTSIndoorLoc data set, the floor labels are in the range \textit{[-3,13]}, and we will move them to the range \textit{[0,16]}

\subsection{Universal Graph Construction - Branch 1}
\label{subsec:2}
After data preprocessing, we begin the branch 1 mentioned in framework overview. To produce prediction results for validation/online data using GNN, we propose a universal and robust graph construction strategy. The construction of graph structures is completely based on preprocessed RSSI features of each receiving point (no task-directed in comparison with Section~\ref{subsec:5}), and strictly abides by the rules of inductive learning graph structure to ensure sufficient generalization ability of trained model, namely:

\hangindent=2.2em
\hangafter=1
1) Graph construction produces three types of graphs: training graph, validation graph and online graph. All nodes in these graphs are receiving points, and no AP in graphs.

\hangindent=2.2em
\hangafter=1
2) In the training phase, training graph structure only contains the training nodes and edges between them, no other nodes or edges shown in the graph.

\hangindent=2.2em
\hangafter=1
3) In the online phase, online graph structure contains training nodes, online nodes, and the corresponding edges. However, the connection between two online nodes is prohibited. The same logic applies to the validation graph.

Table~\ref{tab:1} demonstrates the rules, and all edges are bidirectional. In our inductive learning framework, both the validation and online graphs include the training nodes. This design reflects the real-world deployment scenarios, in which the system must handle individual input nodes during online positioning. In such cases, neighborhood information can only be retrieved from existing training nodes. Therefore, the validation and online graphs are constructed by connecting validation or online nodes exclusively to the training nodes, while internal edges among validation–validation or online–online nodes are strictly prohibited. This ensures that the model operates under consistent inductive conditions and accurately simulates the online positioning process.

We must emphasize that, during validation and online testing, training nodes are included in the graphs solely as neighbors for feature aggregation and do not participate in the computation of validation or test errors.

\begin{table}
\caption{Inductive Learning Graph Structure}
\centering
\begin{tabular}{lcc}
\Xhline{1pt}
\textbf{Graph} & \textbf{Allowed Node Type} & \textbf{Allowed Edge Type} \\
\hline
Train & Train & Train-Train \\
Validate & Train, Validate & Train-Train, Train-Validate \\
Online & Train, online & Train-Train, Train-online \\
\Xhline{1pt}
\end{tabular}
\label{tab:1}
\end{table}

The detailed graph construction method includes the following, and this method also forms the basis for the subsequent task-directed multi-graph construction, which will be described in Section~\ref{subsec:5}.

\subsubsection{Sampling Point Aggregation}
In most Wi-Fi fingerprinting scenarios, all data records are receiving points; therefore, we consider each record as a node. However, there is a problem: fingerprint data usually contains multiple records from one location (sampling point). If the K-nearest neighbors method is directly used to find the nearest neighbors in RSSI space for edge connection, there is a great probability that a sampling point with multiple nodes will have edges connecting between each node internally, which is not helpful for subsequent graph representation learning.

\begin{figure}
    \centering
    \includegraphics[width=\linewidth]{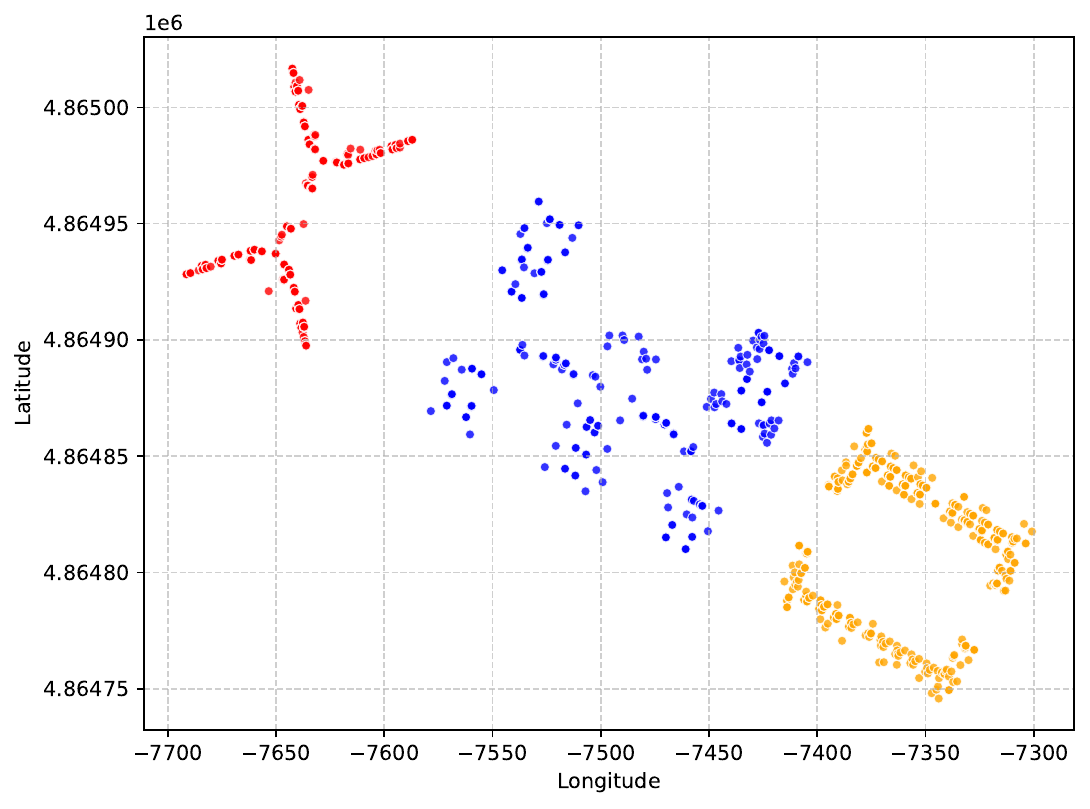}
    \caption{Sampling points of UJIIndoorLoc dataset.}
\label{fig:4.2.1}
\end{figure}

Therefore, it is necessary to prohibit a node from connecting to other nodes that are from its own sampling point, and sampling point aggregation is conducted. It should be emphasized that the aggregation strategy here is only used for training nodes, and online nodes do not have this problem due to the rules shown in Table~\ref{tab:1}. The detail of sampling point aggregation is described in Algorithm~\ref{alg:4.2.1}. As an example, the UJIIndoorLoc dataset’s training data contains 933 sampling points, as shown in Fig.~\ref{fig:4.2.1}.

\begin{algorithm}
\caption{Sampling Point Aggregation Algorithm}
\begin{algorithmic}[1]
\Require Training data $\mathcal{D}$ containing RSSI, position label
\Ensure Training data $\mathcal{D}$ with Sampling Point ID (SPID), Sampling point set $\mathcal{S}$ (aggregated RSSI, position label, SPID)

\State \textbf{Initialization:} $\mathcal{S} \gets \emptyset$
\For{each record $d \in \mathcal{D}$}
    \State Extract position label $p_d$
\EndFor
\State Sort all position labels $\{p_d\}$ to form sampling point set $\mathcal{S}$ 
\State Assign a unique SPID to each sampling point in $\mathcal{S}$
\For{each record $d \in \mathcal{D}$}
    \State Assign the corresponding SPID to $d$
\EndFor
\For{each sampling point $s \in \mathcal{S}$}
    \State Collect all records $d$ associated with $s$
    \State Compute the mean RSSI of these records
    \State Assign the mean RSSI to $s$
\EndFor
\State \Return $\mathcal{D}$, $\mathcal{S}$
\end{algorithmic}
\label{alg:4.2.1}
\end{algorithm}

\begin{algorithm}
\caption{Neighbor Searching Algorithm}
\begin{algorithmic}[1]
\Require Sampling Point Set $\mathcal{S}$ (RSSI, SPID), Training Data $\mathcal{D}$ (RSSI, SPID), Target Node Set $\mathcal{V}$ (RSSI, SPID), KNN parameter $K$, KNN parameter $N$
\Ensure Neighbor sets for each target record

\State \textbf{Initialization:} Fit KNN search on the RSSI of $\mathcal{S}$

\For{each record $r \in \mathcal{V}$}
    \State Search $K+1$ nearest neighbors of $r$ in $\mathcal{S}$ using the KNN model
    \If{$r$ has a valid SPID}
        \If{any neighbor has the same SPID as $r$}
            \State Remove that neighbor from the neighbor list
        \EndIf
    \EndIf
    \State Select the top $K$ most similar neighbors from the remaining list
    \State \textbf{Initialization:} Neighbor set $\mathcal{N}_r \gets \emptyset$
    \For{each selected neighbor $s$}
        \State Find all records in $\mathcal{D}$ whose SPID equals that of $s$
        \State Randomly select up to $N$ records (select all if fewer than $N$)
        \State Add these selected records to $\mathcal{N}_r$
    \EndFor
\EndFor

\State \Return Neighbor sets $\{\mathcal{N}_r\}_{r \in \mathcal{V}}$
\end{algorithmic}
\label{alg:4.2.2}
\end{algorithm}

\subsubsection{KNN Search}
After aggregation, we utilize the K-nearest neighbors~\cite{cover1967nearest} based method to conduct neighbor searching for each target node (Algorithm~\ref{alg:4.2.2}): search K+1 neighbors on sampling point set for the target node; after searching, judge if there is a searched neighbor from this node’s own sampling points, if so, delete it, and finally obtain the top K neighbors; these K neighbors are all sampling points, we randomly select up to N records (nodes) in each sampling point as the final neighbor searching result of this target record (up to K*N).

That is, no matter for the training nodes, verification nodes or online nodes, they will only find the neighbors in the training nodes, and for each training node, it will not find the neighbors from its own sampling point.

\subsubsection{RSSI-based Graph Construction}
\begin{figure}
    \centering
    \includegraphics[width=\linewidth]{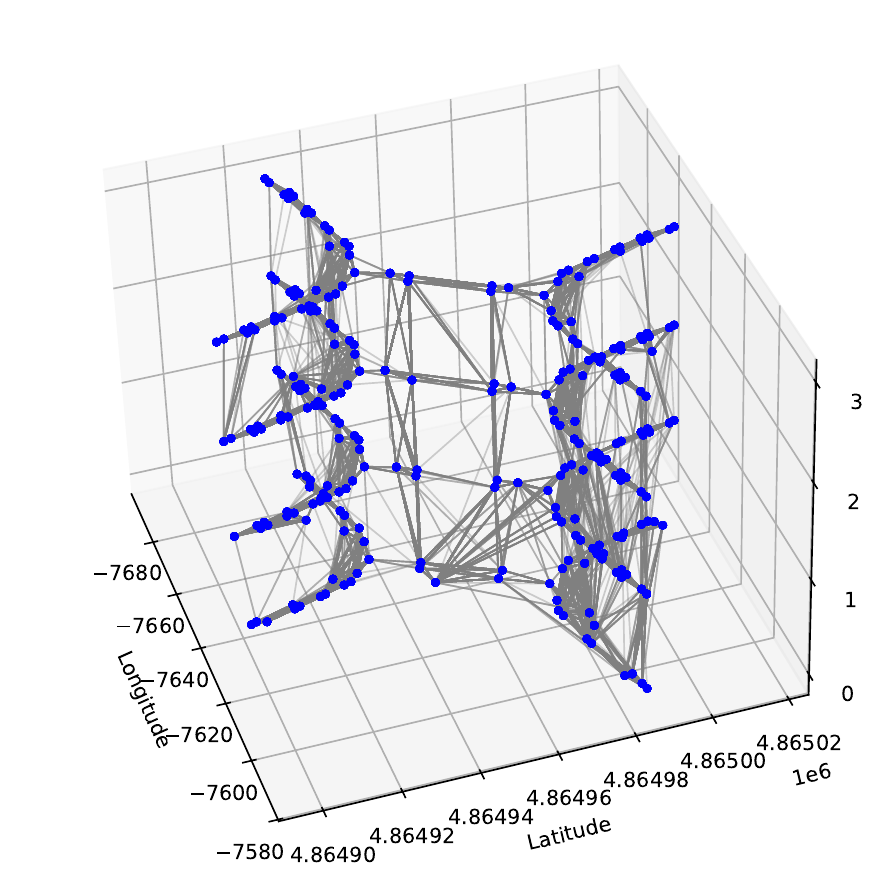}
    \caption{Validation graph of UJIIndoorLoc Building 0.}
    \label{fig:4.2.2}
\end{figure}

After neighbor searching, connect edges between nodes and their neighbors to form the RSSI-based graph structure, which consists of three types of graph: training graph, validation graph and online graph. Fig.~\ref{fig:4.2.2} is an example of validation graph of UJIIndoorLoc building 0.

\subsubsection{KNN Configuration}
When implementing KNN, we use euclidean distance to measure the distance between two normalized RSSI features. The connected edges in all graph structures are bidirectional and have no basic weight. The number of searched neighbors K and the maximum record selection number N of selected sampling points are used as variable parameters of the overall framework.

\subsection{GNN-based Coarse Localization - Branch 1}
\label{subsec:3}
In this part, we propose a light-weight GNN positioning model based on GraphSAGE~\cite{hamilton2017inductive}, which is used for longitude \& latitude regression task and building \& floor classification task. The positioning results of this part will be used in position-based graph construction. The input of this model is preprocessed RSSI features and constructed graph structures. The output is the \textit{[N,2]} longitude \& latitude regression results or \textit{[N,F]}, \textit{[N,B]} floor \& building classification results. The structure of this part is shown in Fig.~\ref{fig:4.3.1}, including GNN module and MLP module. The details of the modules are as follows:

\begin{figure}
    \centering
    \includegraphics[width=\linewidth]{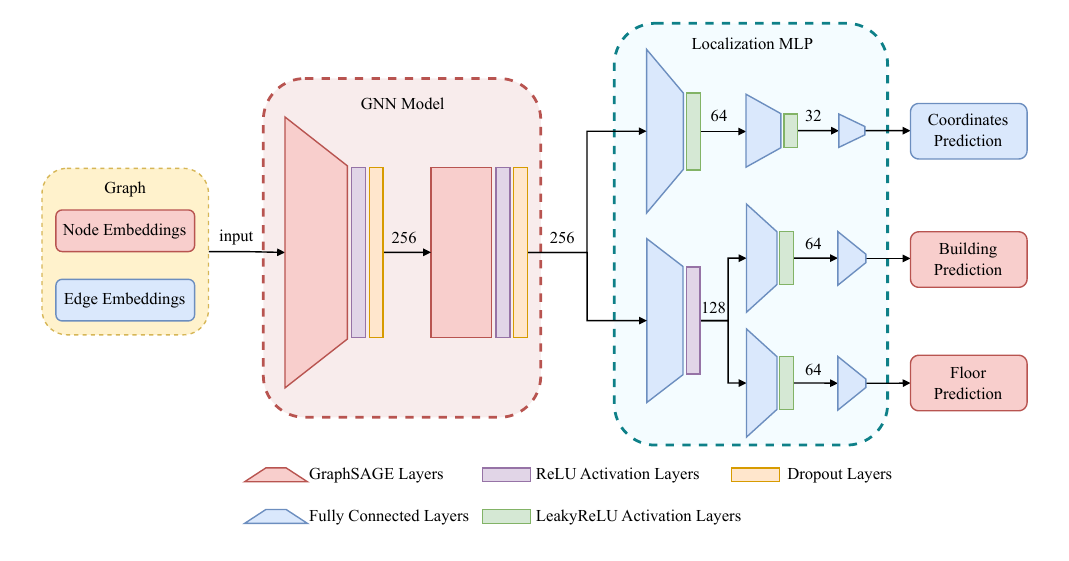}
    \caption{Structure of GNN-based localization.}
    \label{fig:4.3.1}
\end{figure}

% To achieve a good generalization performance, we implement inductive learning here. Inductive learning establishes a general model through training data, which can directly predict any unseen data. In contrast, transductive learning is tailored to a specific test set and cannot be generalized to new test data without retraining. Inductive learning is applicable to the open world hypothesis (test data unknown) and real time predicting, covering the vast majority of supervised learning scenarios. Therefore, in the application of Wi-Fi fingerprint localization, inductive models have obvious advantages over transductive models.

\subsubsection{GNN}
Each GNN layer updates the representation vector of the nodes through sampling and aggregation operations. This process mainly includes three steps: neighbor sampling, information aggregation and node representation update.
For each node, its neighbors are sampled to form a local neighbor set. The purpose of this sampling is to capture local structural and feature information while maintaining scalability on large graphs. GraphSAGE typically adopts uniform sampling, where each node samples the same number of neighbors. After sampling, the feature vectors of the selected neighbors are aggregated to update the target node representation through the aggregation function, which can be mean, max pooling, LSTM or other operations. In our coarse localization method, we use the mean aggregation method:

\begin{equation}
    Aggreg^{(l)}_{mean}(\{\mathbf{h_u^{(l)}}: u \in \mathcal{N}(v)\}) = \frac{1}{|\mathcal{N}(v)|} \sum_{u \in \mathcal{N}(v)} \mathbf{h_u^{(l)}}
\end{equation}
where $\mathcal{N}(v)$ represents the neighbor set of node $v$.

The standard update steps of GraphSAGE are aggregation and nonlinear transformation, and the representation of hidden layer nodes is:

\begin{equation}
    \mathbf{h}_v^{(k)} = 
    \mathbf{W}^{(k)} \cdot 
    \text{aggreg}\left( 
    \left\{ \mathbf{h}_u^{(k-1)} \mid u \in \{v\} \cup \mathcal{N}(v) \right\} 
    \right) 
    \label{formula:6}
\end{equation}

Since the RSSI features have been normalized to [0,1], we utilize \textit{ReLU} activation function on each layer of GNN, and the node representation is:

\begin{equation}
\begin{alignedat}{1}
    \mathbf{h}_v^{(k)}=& \\ &
    \hspace{-2.7em} \text{ReLU}\left( 
    \mathbf{W}^{(k)} \cdot 
    \text{aggreg}\left( 
    \left\{ \mathbf{h}_u^{(k-1)} \mid u \in \{v\} \cup \mathcal{N}(v) \right\} 
    \right) 
    \right)
\end{alignedat}
\end{equation}
% \begin{equation}
%     \text{ReLU}(x) = \max(0, x)
% \end{equation}

\subsubsection{MLP}
We use two types of MLP here: regression task MLP and classification task MLP. The regression task MLP has one output head, while classification task MLP has a shared layer and two separate output heads for multi-tasking.

The input of MLP are the nodes’ eigenvectors produced by GNN, that is, the learned node features represented by the graph neural network. Our MLP uses the fully connected layer (FC) and \textit{LeakyReLU} activation function:
% \begin{equation}
%     {{\mathbf{X}}_{\mathrm{FC}}^{(n+1)}}\mathrm{=}{{\mathbf{W}}^{(n)}}{{\mathbf{X}}_{\mathrm{OUT}}^{(n)}}\mathrm{+}{{b}^{(n)}}
% \end{equation}
\begin{equation}
    {{\mathbf{X}}^{(n)}}=\text{LeakyReLU}\left ({{\mathbf{W}}^{(n)}}{{\mathbf{X}}^{(n-1)}}+\mathbf{b}^{(n)}\right )
\end{equation}
% \begin{equation}
%     \text{LeakyReLU}(x) = 
%     \begin{cases} 
%     x & \text{if } x \geq 0 \\
%     \alpha x & \text{if } x < 0 
%     \end{cases}
% \end{equation}
% Where $\alpha$ is 0.01 by default.

MLP uses the feature representation from GNN to carry out the regression task of longitude and latitude coordinates or the classification task of buildings and floors. When used for multi task classification training of buildings and floors, the convergence speed and loss scale on floor and building output heads are obviously different. Therefore, we design a training strategy that can automatically adjust the loss calculation method in the training process, and control which parameters of the model participate in the training phase. Details are given below.

At the beginning of training, we use the mixed loss (formula (\ref{formula:9}) with $\beta=0.1$) to give priority to training floors, and the purpose of retaining small building loss is to prevent the information related to building classification from being diluted. When floor training converges, the GNN and the floor classification head’s parameters are frozen, and only the building classification loss ($\beta = 1$) is used to separately train the building classification head.
\begin{equation}
    Loss_{mix}\mathrm{=}(1-\beta)Loss_{floor}+\beta Loss_{building}
    \label{formula:9}
\end{equation}
The floor classification task converges more slowly than the building classification. A smaller weight on the building loss prevents it from dominating the gradient updates and allows the model to focus more on optimizing the slower-converging floor-level task. Thus, we set $\beta = 0.1$ for floor training.

In the training phase, we utilize two loss factors, mean squared error (MSE) and cross-entropy for coordinate regression ($Loss_{C}$) and building/floor ($Loss_{B/F}$) classification tasks respectively, namely:
\begin{equation}
    Loss_{C}\mathrm{=}\text{MSE}\mathrm{=}\frac{\mathrm{1}}{n}\sum_{i\mathrm{=1}}^{n}{||\mathbf{y}_{i}-\mathbf{\hat{y}}_{i}||^2}
\end{equation}
where $\mathbf{y}_{i}$ is the ground truth, and $\mathbf{\hat{y}}_{i}$ is the prediction results.
\begin{equation}
    Loss_{B/F}\mathrm{=}\text{Cross-Entropy}\mathrm{=-}\frac{\mathrm{1}}{n}\sum_{i\mathrm{=1}}^{n}{\sum_{j\mathrm{=1}}^{k}{{y_{ij}}}}\mathrm{\log{(}}{p_{ij}}\mathrm{)}
\end{equation}
where $y_{ij}$ is the real one-hot encoding label, and $p_{ij}$ is the predicted probabilities. In implementation, cross-entropy calculation function will automatically perform \textit{softmax} on $p_{i,:}$.

This GNN-based localization model can be trained independently, and the training process will not affect other parts of the proposed framework. Finally, a trained GNN localization model will be given.

\subsection{Stacked Feature Encoder - Branch 2}
\label{subsec:4}
\begin{figure}
    \centering
    \includegraphics[width=\linewidth]{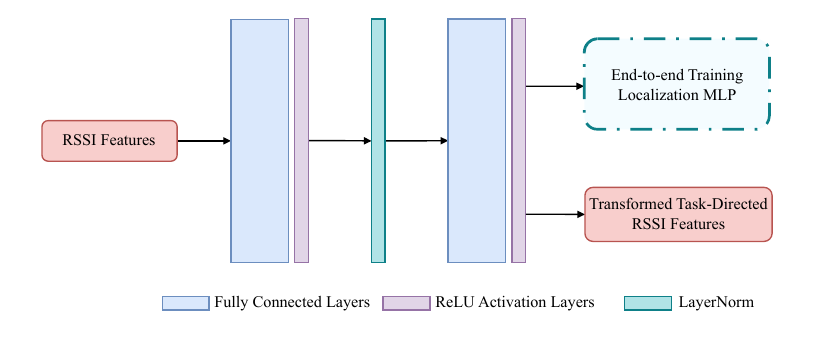}
    \caption{Structure of stacked feature encoder.}
    \label{fig:4.4.1}
\end{figure}

This part is the start of branch 2, which is in parallel with branch 1. We propose a lightweight stacked feature encoder (SFE) model, which is used to rapidly transform the input RSSI features, making it more targeted for certain localization tasks. The model structure is shown in Fig.~\ref{fig:4.4.1}, and details are introduced below.

The inspiration for SFE comes from the stacked auto-encoder (SAE)~\cite{belmannoubi2019stacked}. Stacked auto-encoder is an unsupervised deep learning model, it can be divided into encoder and decoder units, which can automatically discover the hierarchical characteristics of data without labels, provide good initial parameters for deep networks and improve the performance of supervised tasks. Compared with SAE, the SFE designed by us only retains encoding unit for feature transformation (not reconstruction), and we also design an invariant dimension structure, that is, the dimensions of all layers of SFE are consistent. Unlike conventional autoencoders that focus on feature compression and reconstruction, this design aims to perform deep non-linear transformations on the input while preserving the complete feature space. Such isodimensional encoding process enables the network to refine and reparameterize feature distributions in a task-directed manner, allowing the learned representations to better align with the specific objectives of the downstream models. Since the goal is not to reconstruct the input but to transform it into a more discriminative and task-aligned representation, the decoder and reconstruction loss become unnecessary. The encoding layer uses the fully connected layer with \textit{ReLU}:
\begin{equation}
    \mathbf{X}_{\text{RSSI}}^{(n)}=\text{ReLU}\left({\mathbf{W}^{(n)}}{\mathbf{X}_{\text{RSSI}}^{(n-1)}}+\mathbf{b}^{(n)}\right)
\end{equation}
We also add L1 regularization loss to limit the number of non-zero features since the original RSSI features are usually very sparse:
% \begin{equation}
%     \mathbf{x}_{\text{RSSI}}^{(n+1)}=\text{ReLU}\left(\mathbf{X}_{\text{FC}}^{(n+1)}\right)=\text{max}\left(0, \mathbf{X}_{\text{FC}}^{(n+1)}\right)
% \end{equation}

\begin{equation}
    \mathcal{L}_{L1}={\lambda_{L1}}\sum_{i=1}^{d}|w_i|
\end{equation}

% \begin{equation}
%     Loss=Loss_p+{{\mathrm{\mathcal{L}}}_{L\mathrm{1}}}
% \end{equation}
% Where $Loss_p$ is the positioning loss of MSE for coordinates or Cross-Entropy for floors.

In the training process, we cannot know the importance of each feature in advance as well as the ideal result of feature transformation, which makes us lack intermediate supervision for encoder. Therefore, we add an MLP module, which is used to produce localization results to assist the training process. This is a modularized training method, which can reduce the difficulty of overall training, improve efficiency, and enhance the controllability of the model.

\subsection{Task-Directed Multi-Graph Construction - Branch 1\&2}
\label{subsec:5}
This part aims to generate multiple task-directed graph structures for the heterogeneous GNN model in the next part. The inputs here are from branch 1 (Section~\ref{subsec:2} and \ref{subsec:3}) and branch 2 (Section~\ref{subsec:4}), including: true position labels of training data, predicted position labels of validation or online data, transformed RSSI features of all data, preprocessed RSSI features of all data. Position labels include coordinates and floors.

Here we consider two tasks, coordinates regression task and floor classification task. The building classification task is relatively simple, and coarse localization is capable enough to give an accurate prediction. Therefore, the final building results will be produced by the GNN-based coarse localization.

At the beginning, in order to ensure the consistency and compatibility of features after graph representation learning using different graphs, the RSSI features of each node are preprocessed RSSI features; transformed features are only used to conduct graph construction. This step is essential to prevent the two parallel GNN structures in Section~\ref{subsec:6} from producing excessively different node representations, which could lead to failure in the subsequent fusion process.

\subsubsection{Coordinates Regression Task Graphs}
The coordinates regression task graphs focus on coordinate related information, including two graph structures: RSSI-based graph and position-based graph.

RSSI-based graph is constructed using coordinate-task-directed transformed RSSI features. In short, we call it RSSI graph, and the edges of it RSSI edge. The SFE is trained completely on coordinate loss, and is then used to transform RSSI features. After obtaining transformed features, we repeat the graph construction procedure of Section~\ref{subsec:2}.

The position-based graph comes directly from the Euclidean distance calculated using the coordinate labels mentioned above (true and predicted). In short, we call it Pos graph, and the edges of it Pos edge. Pos graph construction strategy is basically the same as that of Section~\ref{subsec:2}. The main difference is that the RSSI features are replaced by coordinates, that is, KNN method searches neighbors in latitude and longitude space instead of RSSI feature space. After searching and selecting, three graphs including train, validate and online graphs will be given.

To effectively transfer these graphs, three heterogeneous graphs (train, val, online) will be generated, each of them includes the information in Table~\ref{tab:2}.

\begin{table}
\caption{Heterogeneous Graph Structure}
\centering
\begin{tabular}{cc}
\Xhline{1pt}
\textbf{Graph Element} & \textbf{Content} \\
\hline
Node Feature & Normalized RSSI \\
RSSI Edge & Edge between nodes based on RSSI \\
Pos Edge & Edge between nodes based on position labels \\
Mask & Mask for training data and validation/online data \\
\Xhline{1pt}
\end{tabular}
\label{tab:2}
\end{table}

The purpose of this graph construction method is to increase the richness of graph information through two different graph structures. The longitude and latitude coordinates used by the verification/online nodes in the graph construction are all prediction results, without disclosing real position labels.

\subsubsection{Floor Classification Task Graphs}
Floor classification task graphs include the same graph structures: RSSI-based and position-based graphs. In RSSI graph construction, compared to coordinates, floor information is relatively less in RSSI features, and we do not wish to establish edges between nodes that are on the same floor but are located at a considerable distance from each other, like building 1 floor 3 and building 2 floor 3. Therefore, the SFE used here is trained on coordinate loss and floor loss using multi-output MLP. In Pos graph construction, there is also a similar problem: we cannot connect edges completely based on floor labels. Therefore, we use all position labels, including coordinates and floors (true and predicted), to construct the graph. The rest procedure is the same as the coordinates regression task graphs, and multiple floor classification task-directed graphs will be given.

\begin{figure*}
    \centering
    \includegraphics[width=\linewidth]{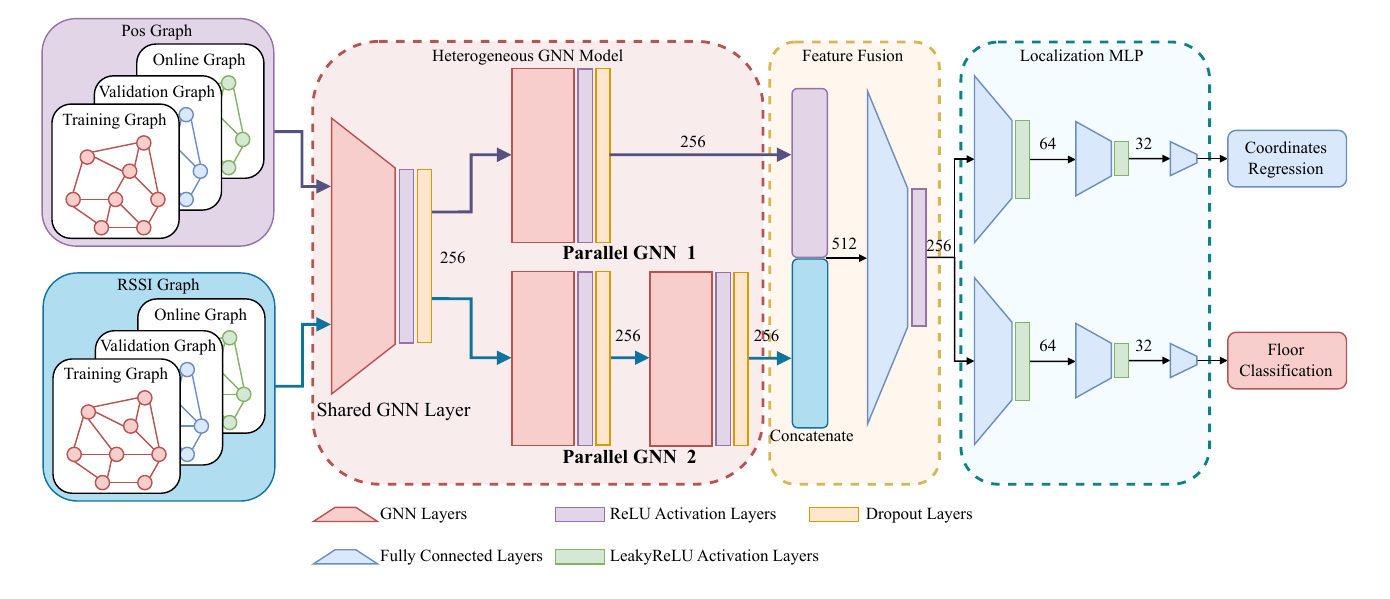}
    \caption{Structure of heterogeneous GNN based localization.}
    \label{fig:4.6.1}
\end{figure*}

\subsection{Heterogeneous GNN based localization - Branch 3}
\label{subsec:6}
In this part, we propose an inductive localization model based on a heterogeneous GNN structure, which is used to output the final localization results. The input of this model is the preprocessed RSSI features from Section~\ref{subsec:1} and the heterogeneous graphs from Section~\ref{subsec:5}, and the outputs are the \textit{[N,2]} latitude and longitude and \textit{[N,F]} floor results. With building results from Section~\ref{subsec:3}, final localization results of overall framework will be produced.

This MG-HGNN model uses GNN modules that can be replaced. This modular design allows the overall framework to serve as a wrapper architecture, where different inductive GNN models (e. g., GraphSAGE, GAT) can be plugged in without altering the overall pipeline.

In this model, the heterogeneous graphs from Section~\ref{subsec:5} will be processed through a shared GNN and parallel GNNs. Then, the feature representations will be concatenated and processed through a fusion layer, and finally, the prediction results will be given by MLP. The structure of this model is shown in Fig.~\ref{fig:4.6.1}, mainly including heterogeneous GNN, feature fusion and localization.

\subsubsection{Heterogeneous GNN}
The structure of the proposed heterogeneous GNN model can be divided into shared layer and parallel layers. The shared layer is the input layer. Its work is to conduct aggregate and nonlinear transform using the input node features and edges of each graph, providing shared parameters for the overall model and preventing the representation difference of parallel layers from being too big to conduct fusion. It can also efficiently reduce the amount of model parameters. For shared GNN layer, we adopt GraphSAGE with mean aggregation to ensure universality and robustness.

The parallel layers are composed of two GNN modules whose parameters are not shared. One is responsible for processing RSSI graphs, and the other is responsible for Pos graphs. Due to the potentially more complex neighborhood connectivity in RSSI graphs, we adopt an additional GNN layer in parallel GNN 2 to better capture high-order interactions among nodes. Finally, two types of representations will be formed. The internal modules of the proposed heterogeneous GNN are replaceable. The parallel GNN modules can adopt inductive GNNs such as GraphSAGE or GAT as required. We can also increase or reduce the number of parallel modules according to the number of graph types. By default, we utilize GraphSAGE with mean aggregation for its superior inductive learning capability, and we will also perform experiments with different GNNs, which will be demonstrated in the experimental section.

The data flow of this model is as follows. Since the internal GNN modules are replaceable, we use $\text{GNN}_{\text{Shared}}$, $\text{GNN}_{\text{Parallel 1}}$ and $\text{GNN}_{\text{Parallel 2}}$ to represent the function of each GNN module, $X$ and $\mathcal{E}$ to represent the node features and edge index of the graph:
\begin{equation}
\left\{
\begin{aligned}
    X'_{\text{Pos Graph}} &= \text{GNN}_{\text{Shared}} \left(X_{\text{Pos Graph}},\mathcal{E}_{\text{Pos Graph}} \right) \\
    X'_{\text{RSSI Graph}} &= \text{GNN}_{\text{Shared}} \left(X_{\text{RSSI Graph}},\mathcal{E}_{\text{RSSI Graph}} \right)
\end{aligned}
\right.
\label{formula:14}
\end{equation}

\begin{equation}
    X_{\text{Parallel 1}} = \text{GNN}_{\text{Parallel 1}} \left(X'_{\text{Pos Graph}}, \mathcal{E}_{\text{Pos Graph}} \right)
\end{equation}
\begin{equation}
    X_{\text{Parallel 2}} = \text{GNN}_{\text{Parallel 2}} \left(X'_{\text{RSSI Graph}}, \mathcal{E}_{\text{RSSI Graph}} \right)
\end{equation}

\subsubsection{Feature Fusion and Localization}
For the output of parallel GNNs, we adopt a fusion method: concatenate the output features, then fuse the features with a fully connected layer to form the final output features of the heterogeneous GNN model (\ref{formula:17}).

\begin{equation}
    \begin{alignedat}{1}
    \mathbf{X}_{\text{Fusion}}=& \\ &
    \hspace{-3.8em} \text{ReLU}\left(\mathbf{W}_{\text{Fusion}}\left(\text{concat}\left(\mathbf{X}_{\text{Parallel 1}}, \mathbf{X}_{\text{Parallel 2}}\right)\right) + \mathbf{b}_{\text{Fusion}} \right)
    \end{alignedat}
    \label{formula:17}
\end{equation}

% \begin{equation}
%     \mathbf{X}_{\text{OUT}}^{(n+1)}=\text{LeakyReLU}\left(\mathbf{W}^{n}\mathbf{X}_{\text{OUT}}^{(n)}+b^{n}\right)
% \end{equation}
This feature will be sent to the MLP for final localization, and the MLP uses fully connected layers and the \textit{LeakyReLU} activation function. The prediction tasks carried out by this MLP are floor classification and coordinate regression tasks, which are the same as those mentioned in the GNN-based coarse localization. This part can be regarded as the fine part of a coarse-to-fine localization strategy. However, it should be pointed out that the prediction of coordinates and floors in GNN-based coarse localization is to provide the position estimation for the second type of graph structure: Pos graph. If we do not make the first estimate, we cannot establish this graph structure for the verification data or the online data. Although Sections~\ref{subsec:3} and \ref{subsec:6} face the same localization tasks, the graphs used and the purpose are obviously different. In addition, our proposed MG-HGNN completely allows the input of other types of graph structure, but taking into account the available information in the majority of Wi-Fi RSSI fingerprint data, we utilize these two graph construction strategies of Pos graph and RSSI graph. More importantly, this framework structure can serve as an effective improvement strategy for GNN-based localization models. We will demonstrate the promotion effect of MG-HGNN on localization results in the experimental results and analysis section.

\subsection{Online Adapter}
\label{subsec:7}
This part is an optional module. We observe that, in the UJIIndoorLoc dataset for example, the data collection times of the training and validation/test sets differ significantly, with an interval of four months~\cite{torres2014ujiindoorloc}. Although the feature semantics remain largely consistent, these sets are still relatively independent. This data independence can better evaluate the model’s generalization ability and practical applicability, but inevitably increases the localization difficulty. Therefore, we introduce an optional online adapter(OA) module to mitigate this issue. This module employs a trainable weight set on the input features and is trained using a small amount (1\%–10\%) of online data, while the main localization models remain fixed in a fully trained state, as:
\begin{equation}
    X'_{\text{online}} = \mathbf{W}_{\text{OA}} X_{\text{online}}
    \label{formula:18}
\end{equation}
OA is proposed to relieve the problem of data irrelevance, it needs a part of online data for training. The training process of OA will not affect the parameters of MG-HGNN. The results from it is for supplementary study only and we will not use it for comparison.

\subsection{Model Complexity}
\label{subsec:8}
Although our framework consists several models, each learning model is a lightweight model composed of a small-scale neural network, so the number of overall parameters is still controllable. Moreover, by introducing a modularized training approach, we successfully avoid training a large number of parameters at the same time, further saving the training cost. To provide a fair comparison, here we take a complete coordinate prediction task of UJIIndoorLoc as an example, Table~\ref{tab:3} gives the parameter count of our MG-HGNN, GAT method and Res-GAT method. The GAT and Res-GAT methods here are single-model GNN methods related to existing studies~\cite{wang2024graph},~\cite{zhang2025graphloc}. For a batch of 256, the proposed framework contains approximately 1.76 million parameters and 1377.60 MFLOPs, which is 5.38 MFLOPs for each node in batch. Although our framework contains multiple models, the scale of parameter and FLOP remains comparable with single-model GNN-based methods, indicating no significant increase in model complexity. This moderate parameter size and computational cost also suggest that the proposed framework is suitable for real-time applications and deployment on computation-constrained devices.

\begin{table}
\caption{Comparison of Model Complexity}
\centering
\begin{tabular}{cccc}
\Xhline{1pt}
\textbf{Model} & \textbf{Parameters} & \textbf{FLOPs (batch)} & \textbf{FLOPs (node)} \\
\hline
GNN (Part C)	&	0.41M	&	61.67M	&	0.24M	 \\
SFE (Part D)	&	0.54M	&	138.98M	&	0.54M	 \\
HGNN (Part F)	&	0.81M	&	1176.95M	&	4.60M	 \\
\midrule
MG-HGNN	&	1.76M	&	1377.60M	&	5.38M	 \\
GAT & 0.66M & 1131.21M & 4.42M \\
Res-GAT & 0.60M & 1839.71M & 7.19M \\
\Xhline{1pt}
\end{tabular}
\label{tab:3}
\end{table}

\begin{table}
\caption{Partition of Datasets}
\centering
\begin{tabular}{ccc}
\Xhline{1pt}
\textbf{Dataset} & \textbf{Parameter} & \textbf{Value} \\
\hline
\multirow{6}{*}{UJIIndoorLoc} &
Training Set Size & 18085 \\
& Validation Set Size & 1852 \\
& Test Set Size & 1111 \\
& Number of APs & 520 \\
& Number of Buildings & 3 \\
& Number of Floors & 4 or 5 \\
\hline
\multirow{6}{*}{UTSIndoorLoc} &
Training Set Size & 8216 \\
& Validation Set Size & 892 \\
& Test Set Size & 388 \\
& Number of APs & 589 \\
& Number of Buildings & 1 \\
& Number of Floors & 16 \\
\Xhline{1pt}
\end{tabular}
\label{tab:4}
\end{table}

\section{Experimental Results and Analysis}
\label{sec:experiment}
All experiments were run on Windows10 using i5-12400f CPU and RTX4060 GPU. The implementation process uses Python 3.12 and PyTorch, and the compiler is Pycharm. Section~\ref{subsec:9} is the datasets description which introduces the public datasets we use. Sections~\ref{subsec:10} and \ref{subsec:11} are the parameter and model configuration tuning parts, in which the validation sets are used. Section~\ref{subsec:12} gives the final performance analysis using test sets. In the validation and test graphs, training nodes serve only as neighbors and are excluded from the loss computation.

\begin{table*}
\caption{Performance of Different KNN factors}
\centering
\begin{tabular}{cccccccc}
\Xhline{1pt}
\multicolumn{3}{c}{\textbf{Parameters}} 
& \multicolumn{3}{c}{\textbf{UJIIndoorLoc}} 
& \multicolumn{2}{c}{\textbf{UTSIndoorLoc}} \\ 
\cline{1-8}
\textbf{K} & \textbf{N} & \textbf{Max Neighbors} & \textbf{Building Acc.} & \textbf{Floor Acc.} & \textbf{MLE (m)} & \textbf{Floor Acc.} & \textbf{MLE (m)} \\ 
\Xhline{1pt}
3 & 1 & 3 & 99.46\% & 96.00\% & 8.37 & 99.10\% & 5.06 \\
3	&	2	&	6	&	\textbf{99.57\%}	&	96.22\%	&	8.25 	&	99.10\%	&	5.11 	 \\
3	&	3	&	9	&	\textbf{99.57\%}	&	96.54\%	&	8.52 	&	99.10\%	&	5.11 	 \\
4	&	1	&	4	&	99.51\%	&	96.54\%	&	\textbf{8.04} 	&	\textbf{99.33\%}	&	\textbf{4.89} 	 \\
4	&	2	&	8	&	\textbf{99.57\%}	&	96.33\%	&	8.44 	&	99.10\%	&	4.98 	 \\
4	&	3	&	12	&	\textbf{99.57\%}	&	\textbf{96.76\%}	&	8.50 	&	99.22\%	&	5.04 	 \\
5	&	1	&	5	&	\textbf{99.57\%}	&	96.33\%	&	8.18 	&	\textbf{99.33\%}	&	5.27 	 \\
5	&	2	&	10	&	99.35\%	&	95.73\%	&	8.40 	&	99.10\%	&	5.43 	 \\
5	&	3	&	15	&	99.41\%	&	96.33\%	&	8.32 	&	99.10\%	&	5.22 	 \\
\Xhline{1pt}
\end{tabular}
\label{tab:5}
\end{table*}

\subsection{Datasets Description}
\label{subsec:9}
We use two public Wi-Fi fingerprint indoor localization data sets, UJIIndoorLoc and UTSIndoorLoc.

The UJIIndoorLoc dataset~\cite{torres2014ujiindoorloc} is a multi-building, multi-floor indoor localization database specifically designed for WLAN/Wi-Fi fingerprint-based positioning systems. Covering three buildings at Universitat Jaume I with over 108,000 $\text{m}^2$ across four or five floors, it contains 21,048 instances comprising 19,937 training records and 1,111 validation/test records. Each record includes 529 features: wireless access points (WAP, 520), position labels, timestamp, etc. The UTSIndoorLoc dataset~\cite{song2019novel} serves as another benchmark. Covering 16 floors of the UTS FEIT Building, which spans approximately 44,000 $\text{m}^2$, it contains 9,496 samples collected from 1,840 distinct locations, with 9,108 records allocated for training and 388 records for testing. Each sample contains 589 WAPs and other information collected.

UJIIndoorLoc provides two files: trainingData and validationData, as its article states, trainingData is for training, and validationData is for validating/testing. UTSIndoorLoc also provides two files: UTS\_training and UTS\_test, respectively for training and testing. In short, we call UJIIndoorLoc UJI and UTSIndoorLoc UTS.

Considering the different numbers of sampling points and sampling records (mentioned in Section~\ref{subsec:2}), we perform sampling point aggregation on the training sets of UJI and UTS, then randomly select 10\% from the sampling point sets, and extract the sampling records that belong to selected sampling points as validation sets. The two files: validationData and UTS\_test, are used for testing. This strategy can enhance the evaluation ability of validation sets, being able to assess the generalization capability of models. The reason for the 10\% selection ratio is that compared to training data records, the number of training data sampling points is obviously less, therefore we can only extract a very small amount of them to prevent insufficient training from occurring. The division of datasets is shown in Table~\ref{tab:4}.

In the testing phase, we adopt two evaluation factors, mean location error (MLE) and classification accuracy, where MLE is the average Euclidean distance, i. e. the following formula:
\begin{equation}
    \text{MLE}=\frac{1}{n}\sum_{i=1}^{n}{\sqrt{({x_{r_i}}-{x_{p_i}})^2+({y_{r_i}}-{y_{p_i}})^2}}
\end{equation}
where $(x_{ri}, y_{ri})$ are the real coordinates and $(x_{pi}, y_{pi})$ are the predicted coordinates. Predicted coordinates come from the inverse transform of \textit{Standardscaler} shown in (\ref{formula:20}), and the accuracy is calculated by using multi-dimension classification output.

\begin{equation}
    Y_{original}=Y_{scaled}\cdot \sigma + \mu
    \label{formula:20}
\end{equation}

% \begin{table*}
% \caption{Performance of Proposed Graph Construction Strategy}
% \centering
% \begin{tabular}{ccccccc}
% \Xhline{1pt}
% \multirow{2}{*}{\textbf{Neighbor Searching Method}}
% & \multicolumn{3}{c}{\textbf{UJIIndoorLoc}} 
% & \multicolumn{2}{c}{\textbf{UTSIndoorLoc}} \\ 
% \cline{2-6}
% & \textbf{Building Acc.} & \textbf{Floor Acc.} & \textbf{MLE (m)} & \textbf{Floor Acc.} & \textbf{MLE (m)} \\ 
% \Xhline{1pt}
% KNN with aggregation & 99.51\% & 96.54\% & 8.04 & 99.33\% & 4.89 \\
% KNN without aggregation & 99.51\% & 96.11\% & 8.54 & 99.33\% & 4.98 \\
% \Xhline{1pt}
% \end{tabular}
% \label{tab:7}
% \end{table*}

\subsection{Framework Parameter Optimization}
\label{subsec:10}

\subsubsection{Parameter Selection of Proposed Graph Construction Method}
The parameters of proposed graph construction method (Section~\ref{subsec:2}) contain two KNN factors, the number of searched neighbors K and the maximum record selection number N of each selected sampling point. We will conduct parameter selection using our proposed GNN-based coarse localization method. Table~\ref{tab:5} shows the performance of different graph construction parameters. Results are produced using validation sets; test sets are not exposed.

Through the experiment on validation sets, the classification accuracy of building and floor labels shows only minor variations. This is reasonable since these labels are primarily determined by coarse-grained spatial patterns, and changing the number of neighbors causes limited information change for classification. For the mean location error, we observe that K/N of 4/1 gives the best result of UJI and UTS. Taking experimental results and computational cost into consideration, we adopt the K/N of 4/1 as the parameter setting of proposed graphs construction methods.

% To demonstrate the effect of proposed graph construction strategy on subsequent GNN-based coarse localization, we compare the results of parameter setting 4/1 with KNN searching without sampling point aggregation in Table~\ref{tab:7}. It shows that there is a noticeable improvement in MLE, and the impact on building and floor classification remains limited. It should be pointed out that the validation sets shares the same distribution as the training set, the benefit of this strategy may be underestimated. Its advantage is expected to be more pronounced when evaluated on distributionally different test sets. For example, the training set and test set of UJI were collected four months apart, leading to significant distribution shifts that make the spatial diversity of neighbors more important.

\begin{figure}
    \centering
    \subfloat[]{
        \includegraphics[width=0.8\linewidth]{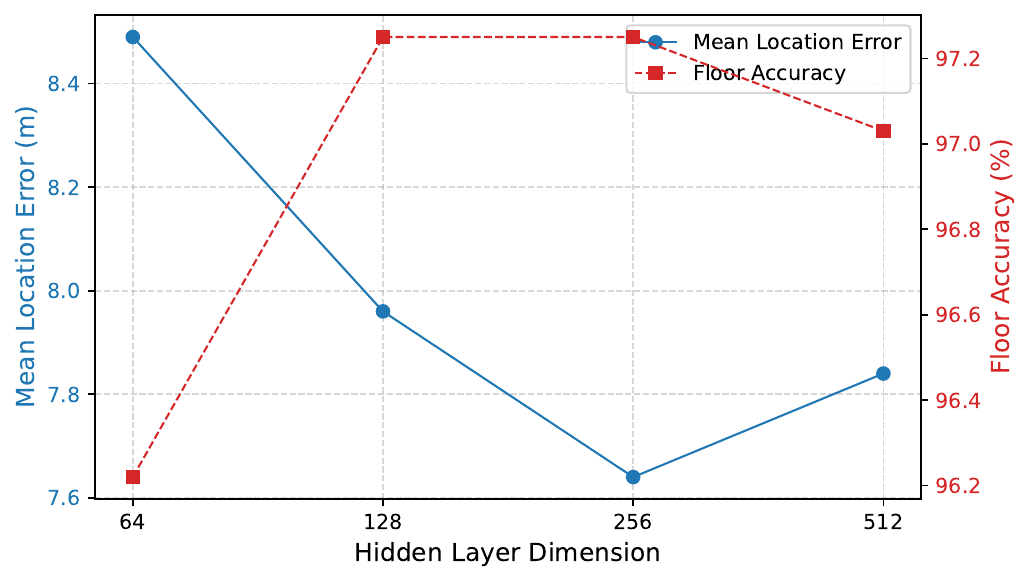}
        \label{fig:5.2.1a}
    }

    \subfloat[]{
        \includegraphics[width=0.8\linewidth]{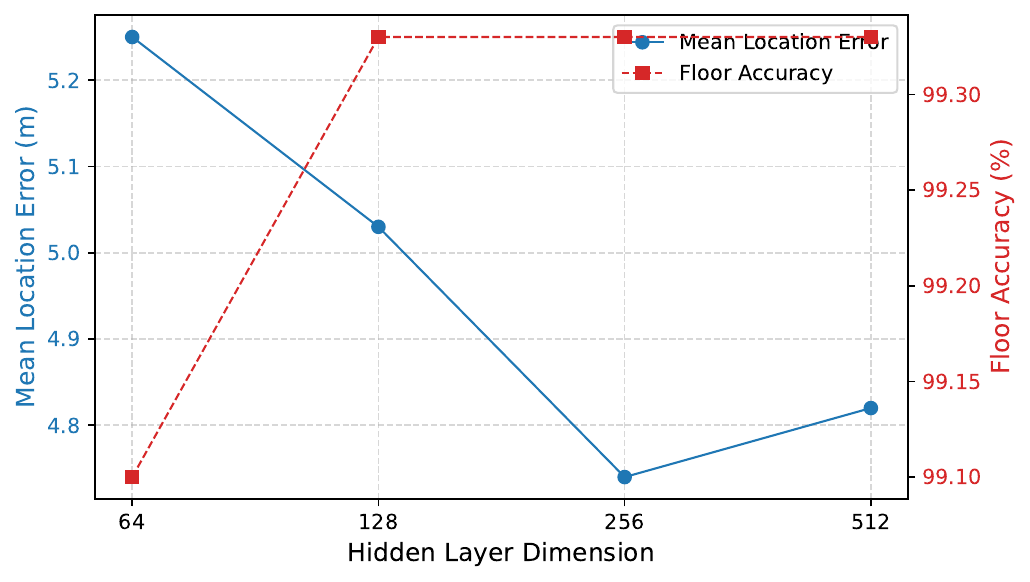}
        \label{fig:5.2.1b}
    }

    \caption{Performance of different hidden layer dimension.\\(a)UJIIndoorLoc. (b)UTSIndoorLoc.}
    \label{fig:5.2.1}
\end{figure}

\begin{figure}
    \centering
    \subfloat[]{
        \includegraphics[width=0.8\linewidth]{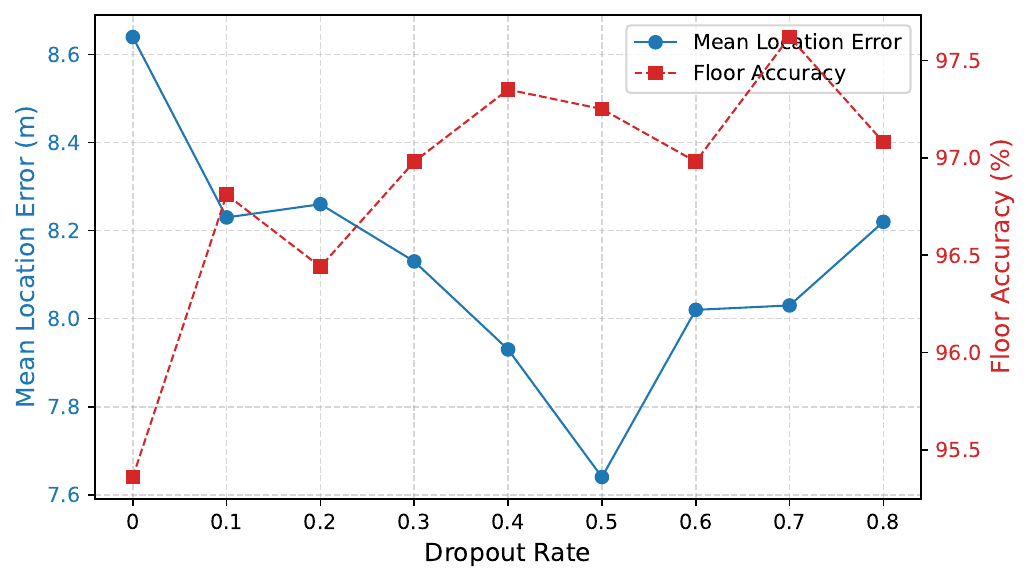}
        \label{fig:5.2.2a}
    }

    \subfloat[]{
        \includegraphics[width=0.8\linewidth]{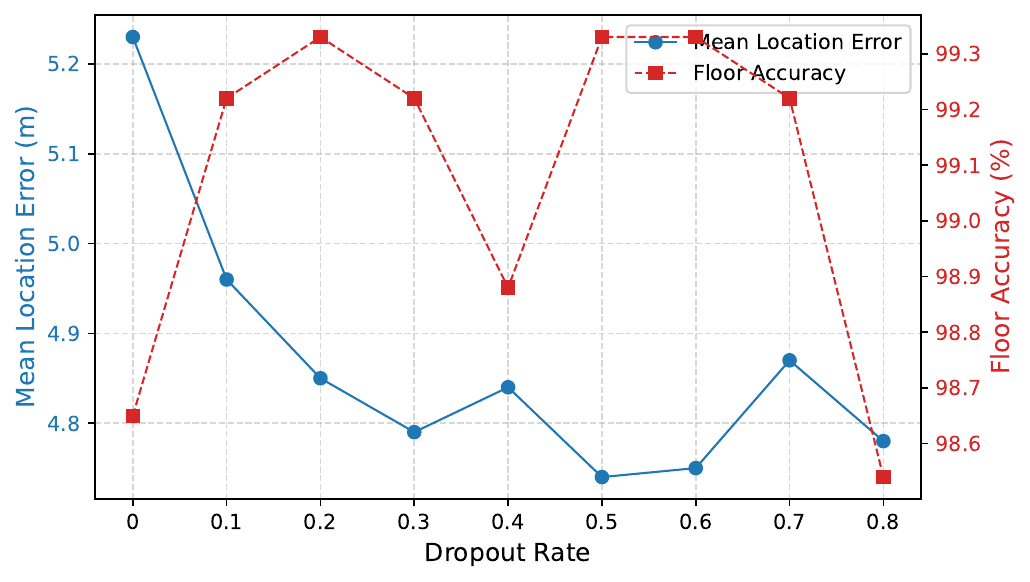}
        \label{fig:5.2.2b}
    }

    \caption{Performance of different Dropout Rate. (a)UJIIndoorLoc.\\(b)UTSIndoorLoc.}
    \label{fig:5.2.2}
\end{figure}

\subsubsection{Parameter Optimization of Heterogeneous GNN}
The main model parameters of our MG-HGNN include the hidden layer dimension and the dropout rate. Different parameters are tested to determine the optimal parameter configuration, and all results are from validation sets using the HGNN model after 100 epochs of training. From Fig.~\ref{fig:5.2.1} and Fig.~\ref{fig:5.2.2}, the mean location error and floor classification accuracy achieve superior results when the hidden layer dimension and dropout factor are 256 and 0.5, thus the optimal parameter setting is determined.

\begin{table}
\caption{Overall Framework Parameters}
\centering
\begin{tabular}{cc}
\Xhline{1pt}
\textbf{Parameter} & \textbf{Value} \\
\hline
Input Dimension	&	520/589	 \\
GNN Hidden Layer Dimension	&	256	 \\
MLP Hidden Layer Dimension	&	64-32	 \\
Output Dimension	&	\makecell{Coordinate: 2 \\ Floor: 5/16 \\ Building: 3/-}	 \\
GNN Activation Function	&	ReLU	 \\
Feature Fusion Activation Function	&	ReLU	 \\
MLP Activation Function	&	LeakyReLU	 \\
Optimizer	&	Adam	 \\
Learning Rate	&	0.0005	 \\
Dropout Factor	&	0.5	 \\
Batch Size	&	256	 \\
KNN parameter K	&	4	 \\
KNN parameter N	&	1	 \\
\Xhline{1pt}
\end{tabular}
\label{tab:6}
\end{table}

\subsubsection{Overall Framework Parameter Configuration}
Based on the experimental results and analysis above, we take the overall framework setting in Table~\ref{tab:6} as our final localization framework for UJIIndoorLoc and UTSIndoorLoc. In training phase, we employ subgraph sampling using \textit{NeighborLoader} to enable mini-batch training, and the batch size is 256.

\begin{table*}
\caption{Performance of MG-HGNN Framework with Different GNNs}
\centering
\begin{tabular}{ccccccc}
\Xhline{1pt}
\multirow{2}{*}{{No.}}
& \multicolumn{2}{c}{\textbf{Method}}
& \multicolumn{2}{c}{\textbf{UJIIndoorLoc}} 
& \multicolumn{2}{c}{\textbf{UTSIndoorLoc}} \\ 
\cline{2-7}
& \textbf{GNN 1} & \textbf{GNN 2} & \textbf{Floor Acc.} & \textbf{MLE (m)} & \textbf{Floor Acc.} & \textbf{MLE (m)} \\ 
\Xhline{1pt}
1	&	GraphSAGE (mean)	&	GraphSAGE (mean)	&	\textbf{97.25\%}	&	\textbf{7.64} 	&	\textbf{99.33\%}	&	\textbf{4.74} 	 \\
2	&	GraphSAGE (max)	&	GraphSAGE (mean)	&	96.76\%	&	8.01 	&	\textbf{99.33\%}	&	4.78 	 \\
3	&	GAT (single attention)	&	GraphSAGE (mean)	&	97.03\%	&	7.93 	&	\textbf{99.33\%}	&	4.75 	 \\
4	&	GAT (multi attention)	&	GraphSAGE (mean)	&	96.54\%	&	7.85 	&	\textbf{99.33\%}	&	4.84 	 \\
5	&	GAT (single attention)	&	GAT (single attention)	&	95.73\%	&	7.83 	&	\textbf{99.33\%}	&	4.80 	 \\
\Xhline{1pt}
\end{tabular}
\label{tab:7}
\end{table*}

\subsection{MG-HGNN with Different GNNs}
\label{subsec:11}
Here we will evaluate the localization performance of our proposed MG-HGNN framework adopting different types of GNNs and parameter settings. For RSSI-based graphs, the RSSI features are inherently noisy and easily affected by device orientation and multipath effects, which can lead to irregular and complex graph structures. Therefore, we consistently use GraphSAGE with mean aggregation on them, which provides a more stable and noise-tolerant feature propagation. For position-based graph, we adopt GraphSAGE with mean and max pooling aggregation, and GAT with single-head and multi-head attention to evaluate which model has the best performance. We also adopt a full GAT configuration for the demonstration of the improvement effect of proposed MG-HGNN framework, which is discussed in the next part.

Table~\ref{tab:7} gives experimental results using the validation sets, in this table, GNN 1 and 2 means the GNN adopted in Parallel GNN 1 and 2 in Fig.~\ref{fig:4.6.1}. From evaluation on validation sets, No.1 MG-HGNN with two GraphSAGEs using mean aggregation gives the best overall results. Therefore, optimal GNN selection for MG-HGNN is using GraphSAGE(mean) in all GNN layers.

\subsection{Performance Analysis and Discussion}
\label{subsec:12}
In this part, we present evaluations on several aspects: the performance improvement achieved by the heterogeneous GNN structure, the effectiveness of the Online Adapter, and the final comparison of positioning accuracy. In addition, an ablation study on is conducted.

\subsubsection{Improvement Effect of MG-HGNN Framework}
Our proposed MG-HGNN framework is capable of supporting different inductive GNN models. To demonstrate the contribution of our framework, we compare our method with original models, highlighting the performance improvements gained from heterogeneous modeling. Evaluation results are produced using the followings methods: MG-HGNN (SAGE) and MG-HGNN (GAT) which is the No.1 and No.5 in Table~\ref{tab:7}, GraphSAGE localization method which uses GraphSAGE with mean aggregation, GAT localization method which uses GAT with single-head attention. GraphSAGE and GAT methods use the homogeneous model structure and the same graph construction strategy in Section~\ref{subsec:2}. Table~\ref{tab:8} gives the comparisons, all results are from test sets and no parameter tuning here or after. Results shows that the MG-HGNN framework effectively improves the performance of GNN models on UJIIndoorLoc and UTSIndoorLoc.

\begin{table*}
\caption{Improvement Effect of MG-HGNN Framework}
\centering
\begin{tabular}{cccccc}
\Xhline{1pt}
\multirow{2}{*}{\textbf{Method}}
& \multicolumn{3}{c}{\textbf{UJIIndoorLoc}} 
& \multicolumn{2}{c}{\textbf{UTSIndoorLoc}} \\ 
\cline{2-6}
& \textbf{Building Acc.} & \textbf{Floor Acc.} & \textbf{MLE (m)} & \textbf{Floor Acc.} & \textbf{MLE (m)} \\ 
\Xhline{1pt}
Proposed MG-HGNN (SAGE)	&	100\%	&	94.24\%	&	7.83 	&	97.16\%	&	6.87 	 \\
GraphSAGE	&	100\%	&	93.79\%	&	8.47 	&	96.13\%	&	7.22 	 \\
\hline
Proposed MG-HGNN (GAT)	&	100\%	&	93.97\%	&	8.43 	&	96.39\%	&	6.89 	 \\
GAT	&	98.92\%	&	88.12\%	&	10.11 	&	93.56\%	&	7.33 	 \\
\Xhline{1pt}
\end{tabular}
\label{tab:8}
\end{table*}

\subsubsection{Performance of Online Adapter}
Our proposed online adapter is used to alleviate the problem of data irrelevance. In the experiment here, 1\%, 5\%, 10\% of the test set is extracted to conduct online training, and the rest of test set is used for testing. MG-HGNN here is trained MG-HGNN (SAGE) with frozen parameters. Table~\ref{tab:9} shows that OA effectively enhances the performance on mean location error, but has little impact on floor classification. Since floor classification depends more on categorical signal patterns rather than continuous spatial alignment, it is reasonable that the online adapter brings little influence.

\begin{table*}
\caption{Performance of Online Adapter}
\centering
\begin{tabular}{cccccc}
\Xhline{1pt}
\multirow{2}{*}{\textbf{Method}}
& \multicolumn{3}{c}{\textbf{UJIIndoorLoc}} 
& \multicolumn{2}{c}{\textbf{UTSIndoorLoc}} \\ 
\cline{2-6}
& \textbf{Building Acc.} & \textbf{Floor Acc.} & \textbf{MLE (m)} & \textbf{Floor Acc.} & \textbf{MLE (m)} \\ 
\Xhline{1pt}
Proposed MG-HGNN	&	100\%	&	94.24\%	&	7.83 	&	97.16\%	&	6.87 	 \\
Proposed MG-HGNN (1\% OA)	&	100\%	&	94.36\%	&	7.81 	&	97.14\%	&	6.83 	 \\
Proposed MG-HGNN (5\% OA)	&	100\%	&	94.50\%	&	7.73 	&	97.02\%	&	6.73 	 \\
Proposed MG-HGNN (10\% OA)	&	100\%	&	94.40\%	&	7.70 	&	97.13\%	&	6.61 	 \\
\Xhline{1pt}
\end{tabular}
\label{tab:9}
\end{table*}

\begin{figure}
    \centering
    \subfloat[]{
        \includegraphics[width=0.8\linewidth]{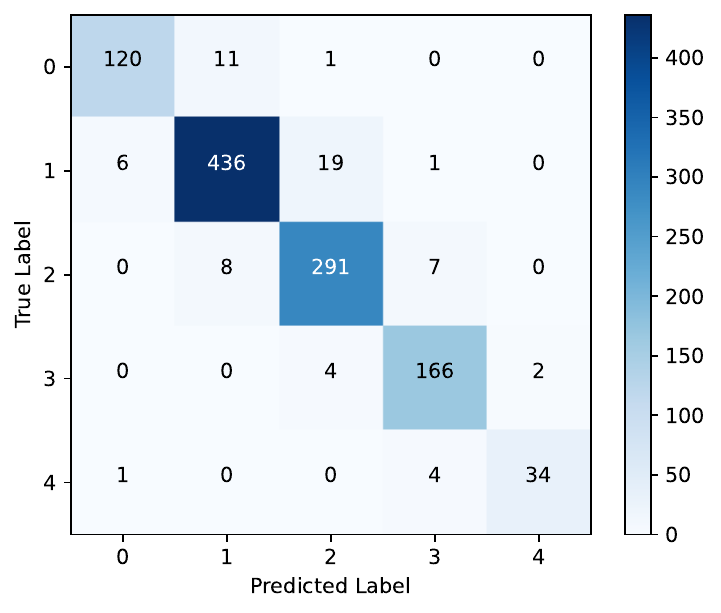}
        \label{fig:5.4.1a}
    }

    \subfloat[]{
        \includegraphics[width=0.8\linewidth]{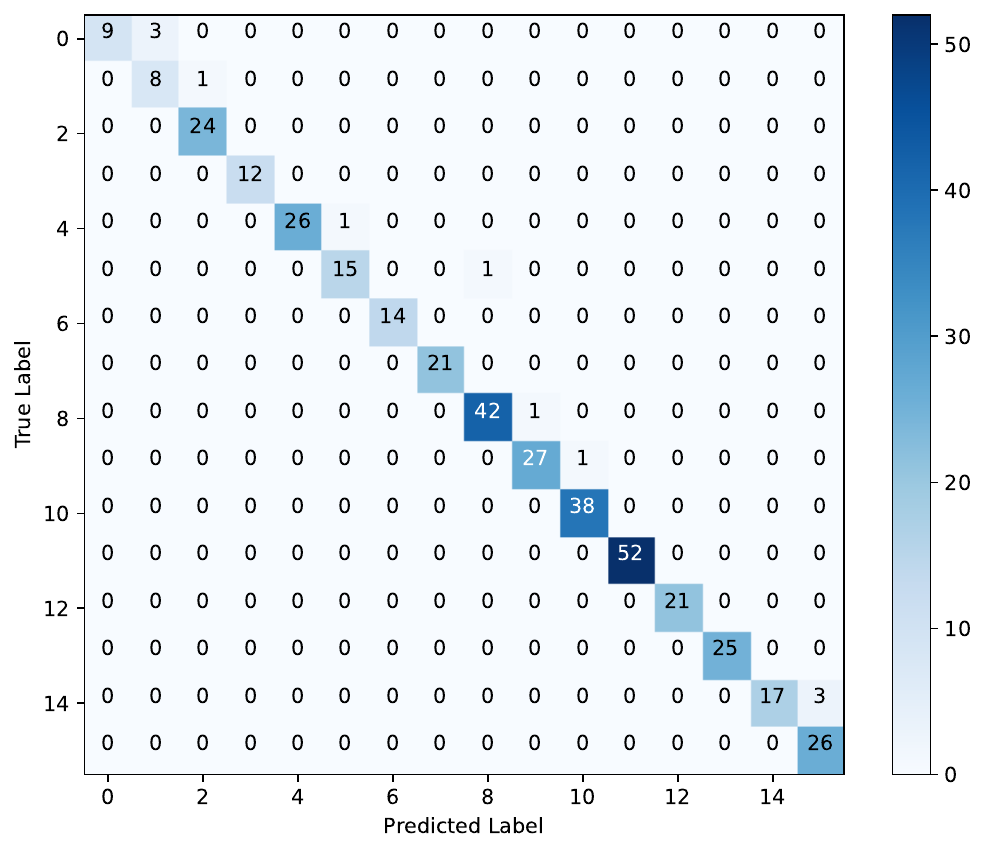}
        \label{fig:5.4.1b}
    }

    \caption{Confusion matrix of floor classification. (a)UJIIndoorLoc.\\ (b)UTSIndoorLoc.}
    \label{fig:5.4.1}
\end{figure}

\begin{figure}
    \centering
    \subfloat[]{
        \includegraphics[width=0.8\linewidth]{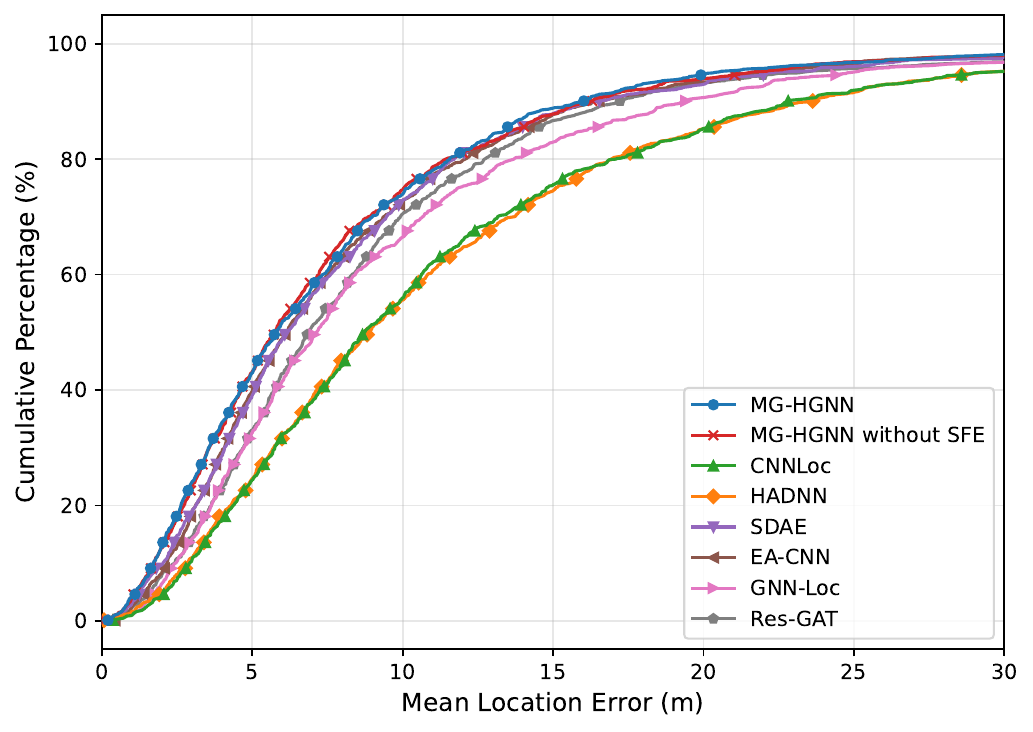}
        \label{fig:5.4.2a}
    }

    \subfloat[]{
        \includegraphics[width=0.8\linewidth]{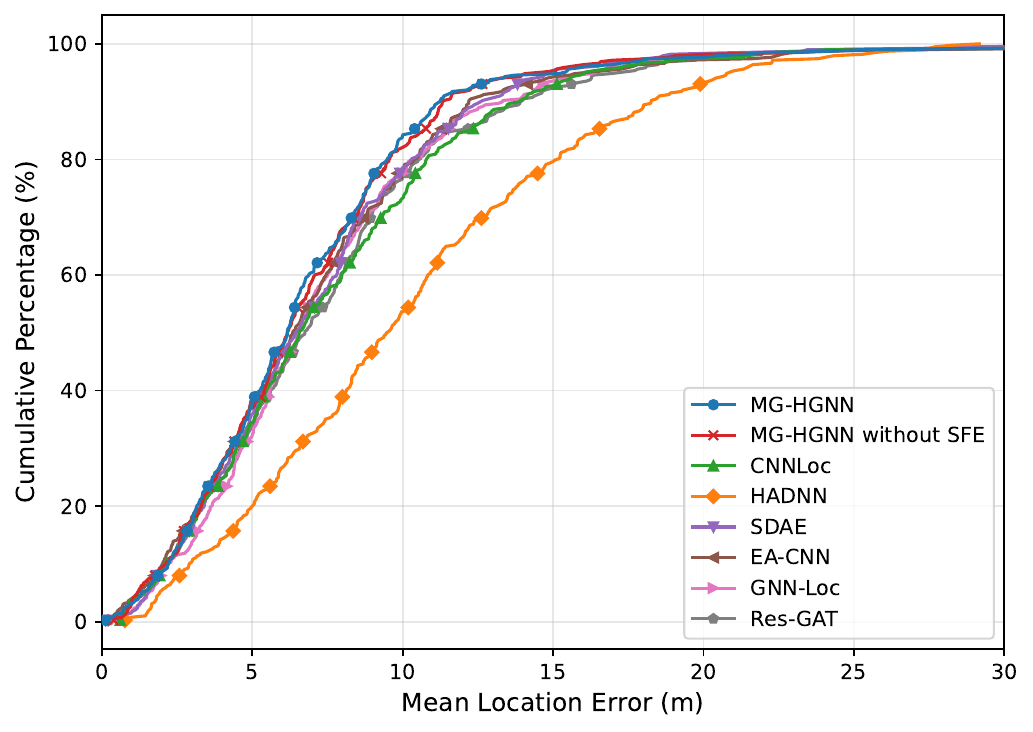}
        \label{fig:5.4.2b}
    }

    \caption{CDF of different localization methods. (a)UJIIndoorLoc.\\ (b)UTSIndoorLoc.}
    \label{fig:5.4.2}
\end{figure}

\begin{table*}
\caption{Positioning Performance of MG-HGNN and Comparison with State-of-the-art Methods}
\centering
\begin{tabular}{cccccc}
\Xhline{1pt}
\multirow{2}{*}{\textbf{Method}}
& \multicolumn{3}{c}{\textbf{UJIIndoorLoc}} 
& \multicolumn{2}{c}{\textbf{UTSIndoorLoc}} \\ 
\cline{2-6}
& \textbf{Building Acc.} & \textbf{Floor Acc.} & \textbf{MLE (m)} & \textbf{Floor Acc.} & \textbf{MLE (m)} \\ 
\Xhline{1pt}
CNNLoc~\cite{song2019novel}	&	99.28\%	&	96.04\%	&	11.78	&	94.59\%	&	7.60	 \\
HADNN~\cite{cha2022hierarchical}	&	\textbf{100\%}	&	93.16\%	&	11.59	&	92.01\%	&	10.24	 \\
SDAE~\cite{zhuang2023robust}	&	99.64\%	&	96.04\%	&	8.37	&	95.36\%	&	7.25	 \\
EA-CNN~\cite{alitaleshi2023ea}	&	\textbf{100\%}	&	\textbf{96.31\%}	&	8.34	&	95.62\%	&	7.21	 \\
GNN-Loc~\cite{wang2024graph}	&	\textbf{100\%}	&	94.15\%	&	9.61	&	\textbf{97.68\%}	&	7.48	 \\
Res-GAT~\cite{zhang2025graphloc}	&	99.64\%	&	91.63\%	&	9.20	&	96.39\%	&	7.66	 \\
\textbf{MG-HGNN}	&	\textbf{100\%}	&	94.24\%	&	\textbf{7.83}	&	97.16\%	&	\textbf{6.87}	 \\
\Xhline{1pt}
\end{tabular}
\label{tab:10}
\end{table*}

\begin{table*}
\caption{Ablation Experiment Comparison}
\centering
\begin{tabular}{cccccc}
\Xhline{1pt}
\multirow{2}{*}{\textbf{Method}}
& \multicolumn{3}{c}{\textbf{UJIIndoorLoc}} 
& \multicolumn{2}{c}{\textbf{UTSIndoorLoc}} \\ 
\cline{2-6}
& \textbf{Building Acc.} & \textbf{Floor Acc.} & \textbf{MLE (m)} & \textbf{Floor Acc.} & \textbf{MLE (m)} \\ 
\Xhline{1pt}
Position-based Graph Only &	\textbf{100\%}	&	93.25\%	&	8.06	&	96.39\%	&	6.96	 \\
RSSI-based Graph Only	&	\textbf{100\%}	&	93.61\%	&	8.25	&	95.88\%	&	7.09	 \\
Shared Layer Only &	\textbf{100\%}	&	92.26\%	&	8.35	&	96.39\%	&	7.35	 \\
Parallel Layers Only &	\textbf{100\%}	&	93.07\%	&	8.30	&	95.62\%	&	7.17	 \\
without SFE &	\textbf{100\%}	&	93.88\%	&	7.92	&	96.65\%	&	6.93	 \\
MG-HGNN	&	\textbf{100\%}	&	\textbf{94.24}\%	&	\textbf{7.83}	&	\textbf{97.16}\%	&	\textbf{6.87}	 \\
\Xhline{1pt}
\end{tabular}
\label{tab:11}
\end{table*}

\subsubsection{Localization Results and Analysis}
To better illustrate the performance on floor classification, the confusion matrices of floor classification results from the proposed MG-HGNN are given in Fig.~\ref{fig:5.4.1}. It shows the distribution of correct and incorrect predictions across all floors. The confusion matrices demonstrate that the proposed framework can accurately classify the majority of floors. Misclassified instances are mostly assigned to adjacent floors, indicating strong classification capability. Nevertheless, some incorrect predictions highlight the need for further optimization strategies. 

To evaluate the performance of proposed MG-HGNN framework comprehensively, we compare and analyze it with several state-of-the-art methods: CNNLoc~\cite{song2019novel}, HADNN~\cite{cha2022hierarchical}, SDAE~\cite{zhuang2023robust}, EA-CNN~\cite{alitaleshi2023ea}, GNN-Loc~\cite{wang2024graph}, Res-GAT~\cite{zhang2025graphloc}, on UJIIndoorLoc and UTSIndoorLoc datasets. Table~\ref{tab:10} shows the comparison results of building and floor classification accuracy, and mean location error. For building classification accuracy of UJI, most existing methods achieve 100\% hit rate. For floor classification accuracy, our proposed MG-HGNN gives the accuracy of 94.24\% and 97.16\% for UJI and UTS respectively. The accuracy of UJI is lower than CNNLoc, SDAE and EA-CNN, but outperforms GNN models like GNN-Loc and Res-GAT, and the accuracy of UTS is higher than most models, secondly only to GNN-Loc. For mean location error, our proposed MG-HGNN achieves 7.83m and 6.87m on UJI and UTS, achieving the best result. This shows that our proposed MG-HGNN framework is capable of producing outstanding localization results.

To further conduct a qualitative evaluation, the CDF curves of mean location error are given in Fig.~\ref{fig:5.4.2}. The CDF curves of our framework outperform those of other methods across most location error thresholds, which further demonstrates the effectiveness and superiority of our framework. This superior performance is attributed to the task-directed multi-graph construction strategy and heterogeneous GNN structure, which significantly enhances the model’s ability to capture spatial and signal information, and improving its targeted capability for specific tasks.

Through the above experimental analysis, the proposed MG-HGNN framework outperforms other models in terms of MLE. However, its performance on floor classification accuracy is not the best. The proposed MG-HGNN effectively reduces the mean location error by leveraging multi-type relational information, demonstrating its capability in capturing complex spatial dependencies. For floor classification, although the framework attempts to capture richer neighborhood information, its advantage over conventional GNNs is not always evident. This can be partly explained by the discrete nature of floor labels and the limited diversity of floor-relevant features at each node, which may reduce the effectiveness multi-type graphs and heterogeneous GNN model. Nevertheless, in the large and complex scenario of UJIIndoorLoc with 1,111 test samples across three buildings, our framework performs better than existing GNN-based approaches. In the smaller and simpler scenario of UTSIndoorLoc with 388 test samples and one building, this effect becomes less pronounced, yet our framework remains competitive and outperforms other non-GNN methods. In general, the results confirm that the proposed framework is effective and provides a practical approach to improving the localization accuracy of existing GNN-based methods, especially in complex environments.

For the ablation study, we conduct experiment on MG-HGNN with different input graphs. The dual inputs of heterogeneous GNN model are position-based graph and RSSI-based graph, to verify this design's effectiveness, we only keep one of these two graphs and input two identical graphs. We also validate the shared-parallel structure of the localization model by only keeping shared layer or parallel layers. In addition, the branch 1 of the MG-HGNN framework is necessary, the overall framework needs it to generate position-based graphs. However, the SFE model of branch 2 is not necessary and RSSI features without transformation can be directly used. Therefore, we add the MG-HGNN without SFE into ablation. The ablation study results presented in Table~\ref{tab:11} demonstrate the rationality and effectiveness of the proposed framework design.

\section{Conclusion And Future Work}
\label{sec:conclusion}
In this paper, we propose a Wi-Fi fingerprint localization framework MG-HGNN based on multi-graph construction and heterogeneous graph neural network. This framework can make full use of RSSI feature information to complete high-precision indoor localization. The multi-type task-directed graph construction method we designed can establish multi-level informative graph structures, and inductive learning can be performed using separate graph structures. The modularized models and the heterogeneous GNN structure can improve the positioning performance while controlling the number of model parameters and training cost. In addition, the online adapter also effectively improves the performance of overall framework. The MG-HGNN has achieved superior performance in comparison with several state-of-the-art methods on two public datasets, UJIIndoorLoc and UTSIndoorLoc, and provides a design paradigm for enhancing existing GNN-based localization methods. 

In future work, we will further optimize the structure of heterogeneous GNN and explore more strategies to build graph structures, so that the positioning model can utilize more multidimensional graph structures with different information. At the same time, we will expand the experiment to more complex and multi-sourcing fingerprint-based positioning data sets, including the data sets we will collect by ourselves. We will further optimize our framework so that it can have stronger generalization capability and better adaptation ability for complicated environments and various sensor data.

\ifCLASSOPTIONcaptionsoff
  \newpage
\fi



%\bibliographystyle{IEEEtran}

\bibliographystyle{IEEEtran}
\bibliography{reference}


% \begin{IEEEbiography}{Yibu Wang}
% \end{IEEEbiography}


%\newpage


%\vfill


\end{document}